\documentclass[lettersize,journal]{IEEEtran}
\usepackage{graphicx}
\usepackage{graphics} 

\usepackage{amssymb}
\usepackage{booktabs}

\usepackage{times}
\usepackage{epsfig}

\usepackage{blindtext}
\usepackage{caption}
\usepackage{amsfonts}
\usepackage[inline]{enumitem}
\usepackage{textcomp}

\usepackage{multirow}
\usepackage{float}
\usepackage{adjustbox}
\usepackage{etoolbox}
\usepackage{dsfont}

\usepackage{lipsum}
\usepackage{stfloats}
\usepackage{multicol}
\usepackage{amsmath,bm}

\usepackage{array}

\usepackage{tabulary}
\usepackage{xcolor}

\usepackage{paralist}

\usepackage[font=footnotesize]{subfig}
\usepackage{ragged2e}

\definecolor{rblue}{rgb}{0,0.5,1}
\definecolor{awesome}{rgb}{1.0, 0.13, 0.32}
\definecolor{hollywoodcerise}{rgb}{0.96, 0.0, 0.63}
\definecolor{lasallegreen}{rgb}{0.03, 0.47, 0.19}
\definecolor{hanpurple}{rgb}{0.32, 0.09, 0.98}
\definecolor{green(pigment)}{rgb}{0.0, 0.65, 0.31}
\usepackage[pagebackref=false,breaklinks=true,colorlinks,bookmarks=false]{hyperref}
\hypersetup{colorlinks=true,
    linkcolor={red},
    citecolor={hanpurple},
    urlcolor={magenta}
}

\usepackage{bbding}

\usepackage{makecell}
\usepackage{stfloats}

\usepackage{icomma}

\def\ie{\emph{i.e.}}
\def\eg{\emph{e.g.}}

\usepackage{xcolor}

\definecolor{generate-color}{RGB}{80,0,80}

\definecolor{mydarkblue}{rgb}{0,0.08,0.45}
\definecolor{myfavblue}{rgb}{0.1176, 0.392, 1.0}
\usepackage{nicefrac}

\usepackage{listings}

\definecolor{dkgreen}{rgb}{0,0.6,0}
\definecolor{gray}{rgb}{0.5,0.5,0.5}
\definecolor{mauve}{rgb}{0.58,0,0.82}
\definecolor{lightgray}{HTML}{DDDDDD}
\usepackage{amssymb}
\usepackage{pifont}
  \definecolor{orange}{HTML}{ff7f0e}
  \definecolor{blue}{HTML}{1f77b4}

\usepackage{longtable}

\usepackage{amsmath}

\newcommand{\repeatcommand}[2]{%
  \ifnum#1>0
    #2%
    \repeatcommand{\numexpr#1-1\relax}{#2}%
  \fi
}

\usepackage{caption}
\usepackage{subcaption}
\usepackage{amsfonts}
\usepackage{color, colortbl}

\definecolor{LightCyan}{rgb}{0.88,1,1}
\begin{document}

\title{RefAtomNet++: Advancing Referring Atomic Video Action Recognition using Multi-Trajectory Semantic Retrieval}

\author{Kunyu Peng$^{*}$ \quad Di Wen \quad Jia Fu \quad Jiamin Wu \quad Kailun Yang$^{*}$ \quad Junwei Zheng \quad Ruiping Liu\\Yufan Chen \quad Yuqian Fu  \quad Danda Pani Paudel \quad Luc Van Gool \quad Rainer Stiefelhagen
\thanks{
\IEEEcompsocthanksitem K. Peng, D. Wen, J. Zheng, R. Liu, Y. Chen, and R. Stiefelhagen are with the Institute for Anthropomatics and Robotics, Karlsruhe Institute of Technology, Germany.
\IEEEcompsocthanksitem J. Fu is with the RISE Research Institutes of Sweden and the KTH Royal Institute of Technology, Sweden.
\IEEEcompsocthanksitem K. Yang is with the School of Artificial Intelligence and Robotics and also with the National Engineering Research Center of Robot Visual Perception and Control Technology, Hunan University, China.
\IEEEcompsocthanksitem J. Wu is with the Chinese University of Hong Kong, Hong Kong, and also with Shanghai AI Lab, China.
\IEEEcompsocthanksitem D. P. Paudel and L. Van Gool are with INSAIT, Sofia University ``St. Kliment Ohridski'', Bulgaria.
\IEEEcompsocthanksitem Y. Fu is with KAUST, Saudi Arabia.
\IEEEcompsocthanksitem $^{*}$Corresponding authors: Kunyu Peng and Kailun Yang.
}%
}

\maketitle

\begin{abstract}
Who is being described, where are they, and what are they doing? Referring Atomic Video Action Recognition (RAVAR) answers these questions jointly by grounding a natural-language reference to a person and recognizing that person's fine-grained atomic actions in complex multi-person videos. Progress in RAVAR is constrained by limited benchmark scale and weak alignment between fine-grained linguistic and scene cues and temporally coherent visual cues. We address both challenges with a new dataset and model. We introduce \texttt{RefAVA++}, a large-scale dataset comprising $2{,}950{,}920$ frames, $75{,}111$ annotated person instances, and $80$ atomic action categories. Its references describe appearance and spatial attributes while deliberately omitting action labels, requiring models to infer actions directly from visual cues. We further propose \texttt{RefAtomNet++}, which models complementary semantics at the holistic-sentence, partial-keyword, and scene-attribute levels. Fine-grained semantic cues retrieve aligned visual tokens across time to construct trajectories, which are aggregated through Mamba-based state-space modeling and fused using multi-hierarchical semantic-aligned cross-attention. This design enables accurate joint person localization and multi-label atomic action recognition. Equipped with the InternVideo2.5 backbone, \texttt{RefAtomNet++} achieves $50.39\%/51.14\%$ mIoU, $59.21\%/59.55\%$ mAP, and $73.89\%/76.59\%$ AUROC on \texttt{RefAVA}, and $49.49\%/50.33\%$ mIoU, $62.33\%/61.90\%$ mAP, and $76.06\%/76.10\%$ AUROC on \texttt{RefAVA++} validation/test sets, respectively. The dataset and code are available at \href{https://github.com/KPeng9510/refAVA2}{\texttt{RefAVA++}}.

\end{abstract}


\section{Introduction}

\begin{figure*}[t]
\centering
\includegraphics[width=\linewidth]{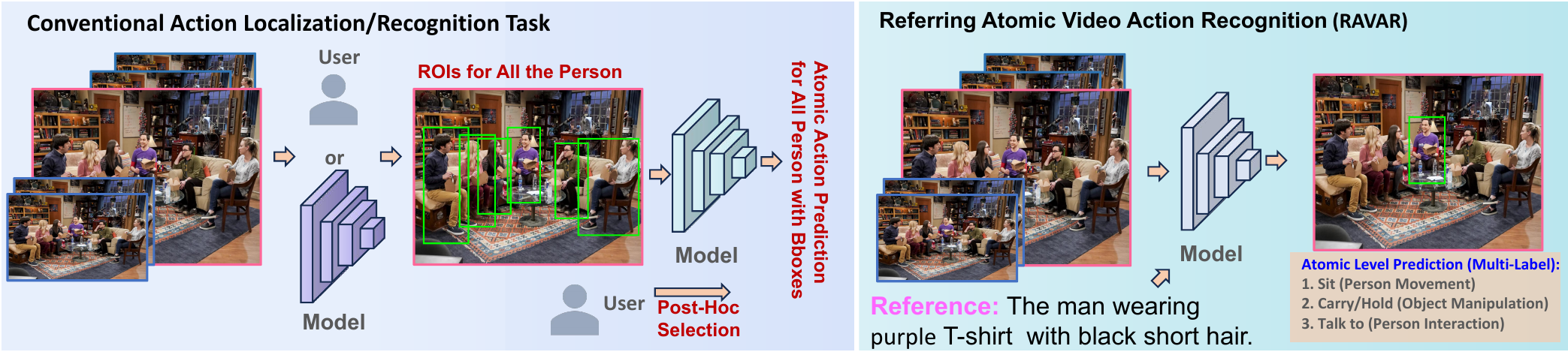}
\caption{Comparison between conventional Atomic Action Localization
(AAL) and Referring Atomic Video Action Recognition (RAVAR).
AAL predicts locations and actions for detected individuals, with an
additional selection step needed to identify a person of interest.
RAVAR uses a natural-language reference to jointly localize the intended
person and recognize their concurrent atomic actions.}
\label{fig:task_comparison}
\end{figure*}

\emph{Who is being described, where are they, and what are they doing?}
Referring Atomic Video Action Recognition (RAVAR)~\cite{peng2024referring}
addresses these questions jointly by taking a video and a natural-language
reference as input and predicting the location and multi-label atomic
actions of the referred person. Unlike conventional referring tasks,
which primarily localize or segment the entity described by
language~\cite{yuan2021instancerefer,wu2023referring,seo2020urvos,dang2023instructdet},
RAVAR also requires fine-grained understanding of that person's behavior
over time. Conventional Atomic Action Localization
(AAL)~\cite{carreira2017quo,gu2018ava} predicts locations and actions for
detected individuals, but identifying a language-specified person
requires an additional selection mechanism. RAVAR incorporates the
reference into localization and action recognition, as illustrated in
Fig.~\ref{fig:task_comparison}.

RAVAR requires visual evidence that supports both identifying the
referred person and recognizing their actions. Global video-text
alignment alone does not explicitly associate individual reference
cues with spatial visual evidence across time.
\texttt{RefAtomNet}~\cite{peng2024referring} addresses this task through
cross-stream fusion of textual, visual, and location-semantic features.
However, it does not explicitly organize visual evidence into separate
temporal sequences conditioned on individual semantic cues.
Attributes such as clothing, relative position, and surrounding
objects can provide complementary information, yet their contributions
may be obscured when aggregated into a holistic representation.
This motivates our central research question:
\emph{How can fine-grained linguistic and scene cues organize visual
evidence across time to improve referred-person localization and
atomic action recognition?}

Investigating this question also requires evaluation resources with
sufficient scale and diversity. As summarized in
Table~\ref{tab:datasets}, existing RAVAR evaluation is constrained by
the limited scale of RefAVA. We introduce \texttt{RefAVA++}, comprising
$32{,}788$ video clips from $231$ movies, $2{,}950{,}920$ frames, and
$75{,}111$ annotated person instances across $80$ atomic action
categories. Each instance is paired with an expert-curated reference
describing identifying attributes while excluding action descriptions.
This design separates language-based target specification from visual
action recognition. Training, validation, and test partitions contain
disjoint movies, supporting evaluation on held-out source movies.
We further evaluate $18$ representative baselines under a
common RAVAR task protocol, covering action localization, video question
answering, video-text retrieval, single-frame grounding, and video
object segmentation on the \texttt{RefAVA++} dataset.

To address this challenge, we propose \texttt{RefAtomNet++}, which uses semantic cues to organize visual evidence across time for joint referred-person localization and atomic action recognition.
Our key idea is to make the evidence associated with each semantic cue an explicit unit of temporal reasoning.
Specifically, at the \emph{partial-keyword} and \emph{scene-attribute} levels, reference words together with detector-derived object semantics and spatial configurations guide the retrieval of relevant spatial visual tokens at each sampled time step.
Tokens retrieved under the same cue are ordered into \emph{semantic trajectories}, exposing the temporal evolution of cue-specific visual evidence.
A complementary \emph{holistic-sentence} pathway retains the complete referential context, preserving information that individual cues may not capture.
Framewise retrieval, however, can yield uneven evidence: a cue may be discriminative in some frames but ambiguous in others.
We therefore employ Mamba~\cite{zhu2024vision} to aggregate these trajectories through selective state-space modeling.
Its recurrent state carries context across sampled frames, while input-dependent updates regulate which information is propagated or discarded.
This provides a mechanism for accumulating informative observations over time, with aggregation complexity linear in trajectory length.
Because compact trajectories can omit spatial context needed for precise localization, we further introduce Multi-Hierarchical Semantic-Aligned Cross-Attention.
Sentence representations and aggregated trajectory features serve as hierarchy-specific queries over dense spatio-temporal video features, reconnecting cue-specific temporal evidence with spatial detail.
Complementary spatial and temporal reasoning branches then predict the referred person's bounding box and multi-label actions.
Together, these operations establish an explicit connection between the semantic cues used to select visual evidence, the temporal context used to interpret it, and the spatial information required to ground the prediction.

Using InternVideo2.5~\cite{wang2025internvideo2} as the backbone, \texttt{RefAtomNet++} achieves $50.39\%/51.14\%$ mIoU, $59.21\%/59.55\%$ mAP, and $73.89\%/76.59\%$ AUROC on \texttt{RefAVA}, and $49.49\%/50.33\%$ mIoU, $62.33\%/61.90\%$ mAP, and $76.06\%/76.10\%$ AUROC on \texttt{RefAVA++} validation/test sets, respectively. 
It also improves action mAP over the standalone backbone on both benchmarks.
Complementary keyframe retrieval analysis provides evidence of above-chance spatial grounding of person-referential cues within the annotated target region.

Our contributions are summarized as follows:
\begin{compactitem}
\item We introduce \texttt{RefAVA++}, a movie-disjoint RAVAR dataset containing $75{,}111$ annotated person instances from $32{,}788$ video clips, with action-free referring expressions and $80$ atomic action categories.

\item We establish a common RAVAR evaluation protocol for $18$ representative baselines spanning five related task families, including recent vision-language and referring-video models.

\item We propose \texttt{RefAtomNet++}, which organizes cue-specific visual evidence into semantic trajectories and integrates temporal state-space aggregation with multi-hierarchical cross-attention, improving referred-person localization and atomic action recognition on both benchmarks.
\end{compactitem}

\section{Related Work}
\label{sec:related_work}

\noindent\textbf{Referring Video and Person-Centric Grounding.}
Referring scene understanding grounds natural-language expressions in visual entities using attributes, contextual reasoning, and cross-modal attention~\cite{shi2018key,ye2021referring,feng2022referring}. 
Its scope has expanded to autonomous driving~\cite{wu2023referring,deruyttere2019talk2car}, and other specialized domains~\cite{bu2022scene,shi2023unsupervised,zhang2023referring}. Video extensions additionally require temporal target association. Representative directions include language-guided video segmentation~\cite{seo2020urvos,khoreva2019video,jin2026interrvos}, referring tracking~\cite{wu2023referring,lin2024echotrack}, transformer-based RVOS~\cite{wu2022referformer}, and spatio-temporal tube grounding~\cite{yang2022tubedetr}. Actor-action segmentation and text-query action localization further connect language with human activities~\cite{gavrilyuk2018actor,mcintosh2020visual,su2023sequence}.

These directions are closely related to RAVAR because they must disambiguate entities and preserve referent identity over time. However, their language query generally specifies the object, event, or action to be localized, and their output is typically a box, mask, track, or temporal interval. 
RAVAR deliberately excludes action descriptions from the reference: language determines \emph{who} is referred to, whereas the video must determine \emph{what that person is doing}. It therefore requires identity grounding and multi-label atomic-action recognition to be solved jointly. 
As summarized in Table~\ref{tab:datasets}, our previous \texttt{RefAVA} benchmark~\cite{peng2024referring} introduced this setting, while \texttt{RefAVA++} increases subject and scene diversity under a movie-disjoint protocol to enable more reliable and generalizable evaluation.

\begin{table*}[!t]
\caption{
An overview of the referring scene understanding datasets, our previously proposed \texttt{RefAVA} dataset~\cite{peng2024referring}, and the proposed \texttt{RefAVA++} dataset in this work.
Our task is Referring Atomic Video Action Recognition (RAVAR), whereas existing benchmarks focus on Referring Object Detection (ROD), Referring Video-based Object Segmentation (RVOS), and Referring Multi-Object Tracking (RMOT).}
\label{tab:datasets}

\scalebox{0.75}{\begin{tabular}{l|llllllllll}
\toprule
\midrule
\textbf{Dataset}       & RefCOCO~\cite{yu2016modeling}                 & RefCOCO+~\cite{yu2016modeling}                & RefCOCOg~\cite{yu2016modeling}                & Talk2Car~\cite{deruyttere2019talk2car} & VID-Sentence~\cite{chen2019weakly} & Ref-DAVIS-2017~\cite{khoreva2019video} & Refer-YV~\cite{zeng2022motr} & Refer-KITTI~\cite{wu2023referring} & RefAVA~\cite{peng2024referring} & \textbf{RefAVA++}   \\
\midrule
\textbf{Task}& ROD & ROD & ROD & ROD & ROD          & RVOS  & RMOT     & RMOT        & RAVAR  & RAVAR   \\
\textbf{N$_{Frames}$}   & 26,711  & 19,992                  & 26,711 & 9,217    & 59,238       & 4,219         & 93,869  & 6,650 & 1,615,140 & 2,950,920\\
\textbf{N$_{Instance}$} & 26,711  & 19,992  & 26,711  & 10,519   & 7,654  & 3,978  & 7,451    & -     & 36,630 &75,111\\
\midrule
\bottomrule
\end{tabular}}

\end{table*}
\noindent\textbf{Atomic Action Recognition and Localization.}
Atomic-action benchmarks such as AVA~\cite{gu2018ava} and HAA500~\cite{chung2021haa500} emphasize fine-grained human movements, with AVA additionally annotating spatially localized concurrent actions in multi-person scenes. Existing methods employ convolutional video encoders such as I3D, SlowFast, and X3D~\cite{carreira2017quo,feichtenhofer2019slowfast,feichtenhofer2020x3d}, or transformer-based architectures such as MViTv2, Hiera, VideoMAE, and end-to-end action transformers~\cite{li2022mvitv2,ryali2023hiera,wang2023videomae,gritsenko2023end}. Person-centric methods further incorporate hierarchical attention, contextual reasoning, pose, or tracking~\cite{pramono2021spatial,ji2019context,rajasegaran2023benefits,wang2023stal}.

Although these methods provide strong action representations, they are generally language-agnostic and predict actions for every detected person, requiring post-hoc target selection. Conversely, referring grounding methods identify a language-specified entity but do not infer an independent set of atomic actions. RAVAR combines these complementary requirements by using language to select the target and person-specific visual evidence to recognize that target's concurrent actions.

\noindent\textbf{Video--Language Retrieval, Question Answering, and Foundation Models.}
Large-scale vision-language pre-training provides transferable representations for video understanding. CLIP and BLIP~\cite{radford2021learning,li2022blip} establish contrastive image-text learning, whereas BLIPv2~\cite{li2023blip} connects frozen encoders through a Querying Transformer. 
Video adaptations include CLIP4Clip~\cite{luo2022clip4clip}, XCLIP~\cite{ma2022x}, ActionCLIP~\cite{wang2023actionclip}, multi-event retrieval~\cite{zhang2023multi}, caption- and tag-enhanced alignment~\cite{wu2023cap4video,chen2023tagging}, and InternVideo~\cite{wang2022internvideo}. 
Video Question Answering methods additionally model hierarchical, causal, and long-range reasoning~\cite{yang2022learning,xiao2023contrastive,gao2023mist,li2022representation,gandhi2022measuring,li2023lavender,bagad2023test}. 
Recent foundation models, including VideoChat~\cite{2023videochat}, Video-LLaMA~\cite{zhang2023video}, Qwen2.5-VL~\cite{ahmed2025qwen}, and InternVideo2.5~\cite{wang2025internvideo2}, offer stronger spatial, temporal, and generative capabilities and can be adapted to target-person action reasoning.

Nevertheless, these models are predominantly optimized for clip-level similarity, action-category alignment, or free-form answer generation. Even multi-granular methods such as XCLIP do not explicitly maintain one referred person's identity across frames while producing joint localization and multi-label action outputs. \texttt{RefAtomNet++} therefore uses foundation models as transferable encoders and supplements them with an explicit referent-conditioned temporal reasoning pathway.

\noindent\textbf{Token Selection and Hierarchical Multimodal Fusion.}
\texttt{RefAtomNet++} is related to token selection and selective attention.
TokenLearner~\cite{ryoo2021tokenlearner} learns compact visual tokens,
while DynamicViT~\cite{rao2021dynamicvit} prunes less informative tokens.
Cross-modal methods emphasize language-relevant
regions~\cite{ye2021referring,liu2021cross,wu2022referformer,li2023blip},
while XCLIP~\cite{ma2022x}, attention
bottlenecks~\cite{nagrani2021attention}, and hierarchical reasoning
models~\cite{le2020hierarchical,jiang2020divide} capture semantic or
temporal interactions.

Our contribution lies in the representation and processing order.
\texttt{RefAtomNet++} retrieves one visual token per keyword or scene-attribute cue at each time step, preserving an explicit \emph{semantic-cue $\times$ time} structure. Mamba~\cite{zhu2024vision} aggregates these cue-indexed sequences with complexity linear in sequence length. The aggregated features and holistic-sentence representation then query the original dense video tokens through a hierarchy-specific cross-attention mechanism.

\section{Benchmark}
\begin{figure*}[t]
    \centering
    \includegraphics[width=1\linewidth]{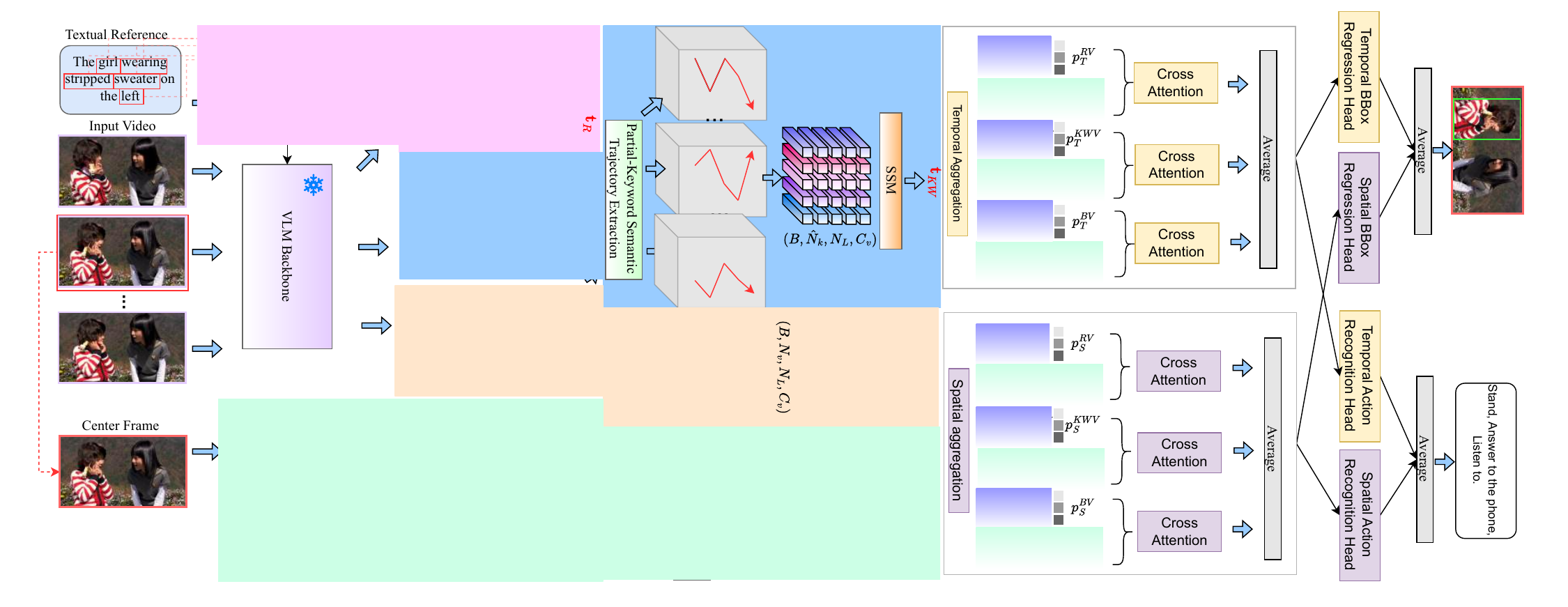}
    \vskip-3ex
    \caption{
    Overview of \texttt{RefAtomNet++}. The model incorporates holistic-sentence, partial-keyword, and scene-attribute representations. Keywords and DETR-derived category and location cues retrieve visual tokens across frames to form semantic trajectories, which separate Mamba layers aggregate. Multi-Hierarchical Semantic-Aligned Cross-Attention integrates these representations with dense video features. Spatial and temporal branches jointly predict the referred person's bounding box and multi-label atomic actions.
    }
    \vskip-1ex
    \label{fig:main_model}
\end{figure*}

\noindent\textbf{Introduction of the RefAVA++ dataset.}
\label{sec:dataset}
To obtain precise textual annotations for the target individuals, $8$ expert annotators were recruited and tasked with providing detailed natural language descriptions based on the key-frame bounding boxes from the AVA dataset~\cite{gu2018ava}. To ensure high annotation quality, a cross-checking process was conducted among annotators, enabling consistency verification and refinement of the textual labels. Discrepant cases were discussed collectively, and only instances that reached clear consensus were retained, while samples with persistent disagreement were discarded. This rigorous protocol ensured both accuracy and linguistic consistency across the dataset, thereby enhancing its reliability for downstream research.

From the AVA dataset, we selected $32,788$ video clips spanning $231$ movies, with an emphasis on clips that exhibit high visual complexity, diverse environments, and multi-person interactions. Each annotated individual was marked at the center frame of every clip, ensuring spatial grounding of the textual reference. In total, the \texttt{RefAVA++} dataset contains $75,111$ annotated instances, partitioned into $48,151$ for training, $16,142$ for validation, and $10,818$ for testing. 

Of particular importance, the test set comprises $4,566$ video clips drawn from $31$ movies distinct from those in the training set ($150$ movies, $21,709$ video clips) and validation set ($50$ movies, $6,513$ video clips). This strict movie-level partitioning enforces challenging generalization conditions and provides a reliable setting for evaluating the model's generalizability in diverse and previously unseen environments.

The textual annotations describe salient visual attributes of the referred subjects, including approximate age, gender, clothing style, physical appearance, and relative position in the frame, \textit{etc}. Notably, action descriptions are intentionally excluded to avoid introducing bias toward predefined action categories and to ensure that models learn to infer actions directly from video evidence rather than textual hints.

Dataset statistics are summarized in Table~\ref{tab:datasets}, where we compare \texttt{RefAVA++} with representative benchmarks in referring scene understanding. The atomic actions span $80$ categories, systematically organized into three groups: Object Manipulation (OM), Person Interactions (PI), and Person Movement (PM). The curated videos capture a wide range of real-world contexts, from everyday interactions to complex multi-person activities.

\noindent\textbf{Baselines.}
\label{sec:baselines}
We largely follow the baseline selection and adaptation protocol 
established in the original RAVAR benchmark~\cite{peng2024referring}, 
covering methods from Atomic Action Localization (AAL), Video Question 
Answering (VQA), Video--Text Retrieval (VTR), Single-Frame (SF) modeling, 
and Video Object Segmentation (VOS). Each method is adapted to jointly 
process the input video and textual reference and to predict both the 
bounding box and multi-label atomic actions of the referred person. 
To incorporate more recent advances, we additionally evaluate 
Qwen2.5-VL-Instruct~\cite{ahmed2025qwen}, 
InternVideo2.5~\cite{wang2025internvideo2}, and 
InterRVOS~\cite{jin2026interrvos}. All previously adopted and newly 
introduced baselines are rerun and evaluated on our newly contributed 
\texttt{RefAVA++} dataset under a unified experimental setting.

\section{RefAtomNet++}

\subsection{Extraction of Visual and Textual Reference Tokens}
\label{sec:token_extraction}

\texttt{RefAtomNet++} is designed to operate with different pretrained video--language feature extractors and the whole architecture is shown in Fig.~\ref{fig:main_model}. In our experiments, we instantiate the framework with either BLIPv2~\cite{li2023blip} or InternVideo2.5~\cite{wang2025internvideo2}. 
Let \(\beta\in\mathcal{B}=\{\mathrm{BLIP\mbox{-}2},\mathrm{InternVideo2.5}\}\) denote the selected backbone. Given an input video \(\mathbf{x}_{V}\) and its textual reference \(\mathbf{x}_{R}\), the backbone-specific feature extractor produces intermediate visual and textual representations, which are subsequently projected into a common \(d\)-dimensional embedding space:
\begin{equation}
\label{eq:1}
\begin{aligned}
\left(\mathbf{h}_V^{(\beta)},\mathbf{h}_R^{(\beta)}\right)
&= \mathcal{E}_{VL}^{(\beta)}(\mathbf{x}_V,\mathbf{x}_R),\\
\mathbf{t}_V &= \mathcal{P}_{VT}^{(\beta)}
       \left(\mathbf{h}_V^{(\beta)}\right),\\
\mathbf{t}_R &= \mathcal{P}_{RT}^{(\beta)}
       \left(\mathbf{h}_R^{(\beta)}\right).        
\end{aligned}
\end{equation}
where \(\mathcal{E}_{VL}^{(\beta)}\) denotes the visual-language feature-extraction pipeline associated with backbone \(\beta\).
The intermediate representations \(\mathbf{h}_{V}^{(\beta)}\) and \(\mathbf{h}_{R}^{(\beta)}\) may have different feature dimensions depending on the selected backbone.
The backbone-specific linear layers \(\mathcal{P}_{VT}^{(\beta)}\) and \(\mathcal{P}_{RT}^{(\beta)}\) map these representations into the shared feature space required by the subsequent semantic-retrieval and trajectory-aggregation modules.
We denote the resulting visual and textual tokens by \(\mathbf{t}_{V}\) and \(\mathbf{t}_{R}\), respectively, and omit the backbone superscript in the remainder of the paper for clarity.

When BLIPv2 is employed, \(\mathcal{E}_{VL}^{(\beta)}\) consists of its pretrained ViT visual encoder, Q-Former, and OPT language model~\cite{li2023blip,zhang2022opt,dosovitskiy2020vit}. 
The Q-Former provides the cross-modal interface between the frame-level visual features and the textual reference. When InternVideo2.5 is employed, we use its native video-language feature-extraction pipeline, including its spatio-temporal visual encoding and adaptive hierarchical token compression, to obtain compact video representations while retaining fine-grained spatial and temporal information. 
Its language component provides the token-level representation of the textual reference.

This unified formulation separates the proposed \texttt{RefAtomNet++} architecture from the internal design of a particular feature extractor. Consequently, the same partial-keyword, scene-attribute, and holistic-sentence reasoning modules can be applied to both backbones without modifying the subsequent semantic-trajectory construction.

\subsection{Multi-Trajectory Semantic-Retrieval Mamba}
\label{sec:SRM-A}
RAVAR requires aligning fine-grained language cues with temporally coherent, person-specific visual evidence. A single holistic-sentence representation can obscure discriminative attributes and scene context, limiting precise grounding and subtle action recognition. We therefore model three complementary semantic levels: holistic-sentence, partial-keyword, and scene-attribute. Fine-grained semantic cues retrieve their nearest visual tokens across frames to form semantic-aligned trajectories, which are aggregated by Mamba. Its selective state-space modeling efficiently captures long-range dependencies and smooth action dynamics while filtering irrelevant evidence, yielding stronger representations for referred-person localization and fine-grained action reasoning. 
The details of the architecture design are described in the following sections.

\noindent\textbf{Partial-Keyword Trajectory Aggregation (PKTA).} Beyond the holistic sentence-level reasoning, partial-keyword-level reasoning captures finer semantic details that can further improve the alignment between semantic and visual tokens at a finer semantic granularity. Given the textual reference input as Eq.~\ref{eq:1_},
\begin{equation}
\label{eq:1_}
    \mathbf{t}_R = \{\mathbf{e}_1, \mathbf{e}_2, \dots, \mathbf{e}_{N_K}\}, \quad
\mathbf{e}_i = \mathcal{V}_{RT}(\mathbf{w}_i) \in \mathbb{R}^d,
\end{equation}
where $N_K$ denotes the total number of words from the given reference sentence, and $\mathbf{e}_i$ indicates the semantic embedding of the $i$-th word, which has dimension $d$. $\mathbf{w}_i$ indicates the $i$-th original word of the textual reference $\mathbf{t}_R$. We first set up a set of stop words, denoted by $\Omega_{stop}$. We then mask out all the stop words (\eg, \textit{the}, \textit{a}, \textit{is}, \textit{are}) existing in the textual reference, as Eq.~\ref{eq:2}.
\begin{equation}
\label{eq:2}
    \hat{\mathbf{t}}_R = \mathbf{t}_R / \Omega_{stop}, 
\end{equation}
where $\hat{\mathbf{t}}_R$ denotes the filtered textual reference. The resultant total number of keywords after filtering is $\hat{N}_K$.

Then, for the semantic embedding of each keyword of $\hat{\mathbf{t}}_R$, we retrieve its nearest spatial visual token for each time step, and thereby obtain the desired trajectory of this keyword along the temporal axis. 
The resultant trajectories are harvested based on all the keywords for a given textual reference. The nearest visual token for each time step can be obtained according to Eq.~\ref{eq:3}.
\begin{equation}
\label{eq:3}
    \mathbf{t}_V^k(l) = \arg\min_{\mathbf{t}\in \mathcal{T}(l)} \; \|\hat{\mathbf{e}}_k - \mathbf{t}\|, 
\quad l = \left[1,\dots,N_L\right], 
\end{equation}
where $\hat{\mathbf{e}}_k$ is the $k$-th keyword embedding of the reference $\hat{\mathbf{t}}_R$, $\mathcal{T}(l)$ is the set of spatial-visual tokens at time step $l$, $\mathbf{t}_V^k(l)$ is the nearest visual token to $\hat{\mathbf{e}}_k$ at time step $l$. 
Next, we obtain the temporal trajectory of each keyword by Eq.~\ref{eq:4}.
\begin{equation}
\label{eq:4}
    \mathcal{T}_k = \{\mathbf{t}_V^k(1), \mathbf{t}_V^k(2), \dots, \mathbf{t}_V^k(N_L)\},
\end{equation}
where $\mathcal{T}_k$ denotes the trajectory of the $k$-th keyword embedding $\hat{\mathbf{e}}_k$, and $N_L$ is the total number of time steps. Finally, we get all the trajectories for all the keywords of the given textual reference by Eq.~\ref{eq:5}.
\begin{equation}
\label{eq:5}
    \mathcal{T}(\hat{\mathbf{t}}_R) = \{\mathcal{T}_1, \mathcal{T}_2, \dots, \mathcal{T}_ {\hat{N}_K}\},
\end{equation}
where $\mathcal{T}(\hat{\mathbf{t}}_R)$ represents the set of trajectories for all the keywords.

We then employ Mamba~\cite{zhu2024vision} to aggregate the trajectories because its selective state-space mechanism efficiently captures long-range temporal dependencies while remaining computationally scalable.
This allows the model to integrate spatial-semantic cues along each trajectory into a compact representation that preserves both local dynamics and global context, as in Eq.~\ref{eq:Mamba_abb}. Unlike attention-based aggregation, which performs discrete pairwise weighting, our Mamba-based state-space modeling can be interpreted as learning a continuous-time dynamical system that filters semantic trajectories through recurrent latent transitions. This inductive bias aligns naturally with human action dynamics, where atomic actions evolve smoothly rather than through abrupt frame-wise changes.
\begin{equation}
\label{eq:Mamba_abb}
    \mathbf{t}_{KW} = \operatorname{Mamba}_{KW}\!\left(\mathcal{T}(\hat{\mathbf{t}}_R)\right),
\end{equation}
where we leverage Mamba~\cite{zhu2024vision} to achieve token aggregation based on the acquired diverse semantic-retrieval trajectories, while the detailed procedure is described in Eq.~\ref{eq:Mamba},
\begin{equation}
\label{eq:Mamba}
\begin{gathered}
    \mathbf{h}_{k}(l) = \mathbf{A}_{KW}(\Delta_l)\,\mathbf{h}_{k}(l-1)
                      + \mathbf{B}_{KW}(\Delta_l)\,\mathbf{t}_V^k(l), \\
    \mathbf{t}_{KW} = \mathbf{C}_{KW}(\Delta_l)\,\mathbf{h}_{k}(l), \\
    \qquad l=\left[1,\dots,N_L\right],\ \ k=\left[1,\dots,\hat{N}_K\right],
\end{gathered}
\end{equation}
where $\mathbf{t}_V^k(l)$ is the selected visual token on trajectory $\mathcal{T}_k$ at time step $l$, $\mathbf{h}_{k}(l)$ is the hidden state of Mamba, and $\mathbf{t}_{KW}$ is the aggregated output.
$\mathbf{A}_{KW}(\cdot),~\mathbf{B}_{KW}(\cdot),~\mathbf{C}_{KW}(\cdot)$ are learnable state-space model kernels, and $\Delta_l$ is the step size, where we use $\Delta_l\!=\!1$ for uniform steps. 
The keyword-semantic-retrieval tokens, \ie, $\mathbf{t}_{KW}$, serve as the input to the query projection in our multi-hierarchical semantic-aligned cross-attention mechanism, which will be introduced later.

\noindent\textbf{Scene-Attribute Trajectory Aggregation (SATA).}
Beyond the partial-keyword-semantic trajectory extraction and token aggregation, we introduce a Scene-Attribute Trajectory Aggregation (SATA) module, designed to explicitly capture contextual cues from the surrounding environment that are often neglected when models focus solely on the referred human or textual information. Concretely, we apply an off-the-shelf object detector to the keyframe (\textit{i.e.}, the center frame) of each video clip, and utilize its outputs to extract both bounding box coordinates and semantic embeddings of the detected object categories. These outputs are then transformed into scene-attribute tokens that encode spatial layouts together with semantic context by concatenation along the feature dimension axis, thereby enriching the representation space with complementary scene-level information.

For object detection, we employ the transformer-based DETR~\cite{carion2020end}, which outputs $N_b$ detection results on the keyframe. The detected instances include both humans and other salient objects with high confidence scores, as formulated in Eq.~\ref{eq_2}. In this formulation, the bounding box coordinates capture the spatial layout of the scene, while the semantic embeddings provide categorical priors. When combined, these elements form scene-attribute-aware token representations that integrate spatial and semantic context for subsequent reasoning.
\begin{equation}
\label{eq_2}
\mathbf{t}_{boxes}, \mathbf{t}_{cats} = \mathcal{V}_{dets}(\mathbf{x}_k),
\end{equation}
where $\mathbf{x}_k$ denotes the keyframe and $\mathcal{V}_{dets}$ is the detection network. Here, $\mathbf{t}_{boxes}\in \mathbb{R}^{N_b \times 4}$ are the detected 2D bounding box coordinates (top-left and bottom-right), where $N_b$ indicates the total number of detected objects. 
$\mathbf{t}_{cats}$ are the predicted categories of the detected regions in textual form. The category labels are subsequently processed by a text encoder (\textit{e.g.}, the textual encoder of BLIPv2~\cite{li2023blip} or InternVideo2.5~\cite{wang2025internvideo2}) to obtain semantic embeddings.

Scene-attribute tokens are particularly beneficial for the RAVAR task. 
Many real-world activities are inherently defined not only by human motion but also by the contextual objects involved (\textit{e.g.}, ``answering the phone'' cannot be interpreted without the presence of a phone). 
Incorporating detected scene attributes enables the model to disambiguate visually similar actions by grounding them in their contextual environment. Moreover, spatial configurations of objects often serve as cues of affordances and plausible human-object interactions (\textit{e.g.}, a nearby chair strongly suggests the possibility of ``sitting''). Finally, scene attributes provide a stable semantic prior: while motion patterns can vary significantly across instances, object co-occurrence remains more consistent, thereby enhancing retrieval robustness in unconstrained, open-domain scenarios.

We encode both semantic and spatial information by concatenating the semantic embeddings with the bounding box coordinates, followed by a linear projection layer to obtain the scene-attribute tokens, \ie, $\mathbf{t}_{BS}$, as defined in Eq.~\ref{eq:ls}.
\begin{equation}
\label{eq:ls}
\mathbf{t}_{BS} = \mathbf{P}_{BS}(\mathrm{Concat}\left[\mathcal{V}_{RT}(\mathbf{t}_{cats}), \mathbf{t}_{boxes}\right]),
\end{equation}
where $\mathbf{P}_{BS}$ is a fully connected layer, $\mathcal{V}_{RT}$ denotes the language feature extraction backbone, and $\mathrm{Concat}$ is the concatenation operator. 
The resulting embeddings $\mathbf{t}_{BS} = \{\mathbf{b}_1, \mathbf{b}_2, ..., \mathbf{b}_{N_b}\}$ are then aligned with spatial-visual tokens across time. Then, for each extracted scene attribute, we retrieve its nearest visual token at each time step to form its corresponding temporal trajectory, as defined by Eq.~\ref{eq:11}.
\begin{equation}
\label{eq:11}
\widetilde{\mathbf{t}}_V^j(l) = \arg\min_{\mathbf{t}\in \mathcal{T}(l)} ||\mathbf{t}_{BS}^j - \mathbf{t}||,
\quad l = \left[1,\dots,N_L\right],
\end{equation}
where $\mathbf{b}_j$ is the $j$-th scene-attribute token in the keyframe, $\mathcal{T}(l)$ denotes the set of spatial-visual tokens at time step $l$, and $\widetilde{\mathbf{t}}_V^j(l)$ is the nearest visual token to $\mathbf{b}_j$ at timestamp $l$ for the $j$-th recognized object. $N_b$ is the total number of detected bounding boxes. 
The temporal trajectory of each detected attribute token is then given by Eq.~\ref{eq:13}.
\begin{equation}
\label{eq:13}
\mathcal{T}_j^{BS} = \{\widetilde{\mathbf{t}}_V^j(1), \widetilde{\mathbf{t}}_V^j(2), \dots, \widetilde{\mathbf{t}}_V^j(N_L)\},
\end{equation}
where $\mathcal{T}_j^{BS}$ denotes the trajectory of the $j$-th scene-attribute token $\mathbf{b}_j$. 
Collectively, the set of trajectories for all detected scene attributes is represented as Eq.~\ref{eq:12}.
\begin{equation}
\label{eq:12}
\mathcal{T}(\mathbf{t}_{BS}) = \{\mathcal{T}_1^{BS}, \mathcal{T}_2^{BS}, \dots, \mathcal{T}_{N_b}^{BS}\},
\end{equation}
where $\mathcal{T}(\mathbf{t}_{BS})$ indicates the trajectory set for all scene attributes.

Finally, we employ a Mamba scanner to aggregate the trajectories $\mathcal{T}(\mathbf{t}_{BS})$ into refined temporal representations as in Eq.~\ref{eq:14}.
\begin{equation}
\label{eq:14}
\hat{\mathbf{t}}_{BS} = \operatorname{Mamba}_{BS}\left(\mathcal{T}(\mathbf{t}_{BS})\right),
\end{equation}
where $\operatorname{Mamba}_{BS}(\cdot)$ is a selective state-space model and $\hat{\mathbf{t}}_{BS}=\{\hat{\mathbf{t}}_{BS}(1),\dots,\hat{\mathbf{t}}_{BS}(N_L)\}$ are the per-timestep aggregated tokens for the scene-attribute-level reasoning.

\noindent\textbf{Holistic-Sentence Token Aggregation (HSTA).} 
Since we adopt a cross-modal feature extractor from the vision-language encoder, the resulting visual tokens incorporate the global semantic context of the reference sentence. Mamba~\cite{zhu2024vision} is well suited to aggregating these tokens because its selective state-space updates accumulate contextual information while adaptively regulating each token's contribution, with computational complexity linear in sequence length. We therefore employ Mamba to obtain holistic sentence-conditioned visual-textual representations, as formulated in Eq.~\ref{eq:15}.
\begin{equation}
\label{eq:15}
\hat{\mathbf{t}}_{V} = \operatorname{Mamba}_{V}\left(\mathbf{t}_V\right),
\end{equation}
where $\mathbf{t}_V$ denotes the video token, which is fused with the whole sentence semantic feature from the reference by Q-Former, $\operatorname{Mamba}_{V}(\cdot)$ is another selective state-space model, $N_v$ indicates the number of visual tokens, and $\hat{\mathbf{t}}_{V}$ are final resultant tokens.

\subsection{Multi-Hierarchical Semantic-Aligned Cross-Attention Mechanism (MHS-CA)}
\label{sec:MHS-CA}
After obtaining the multi-trajectory semantic-retrieval tokens, we introduce a multi-hierarchical semantic-aligned cross-attention mechanism. This mechanism helps selectively integrate visual tokens and align complementary information between the retrieved tokens and the original visual-language tokens, from both spatial and temporal perspectives.
We employ a two-branch architecture that aggregates temporal and spatial information through the proposed multi-hierarchical semantic-aligned cross-attention mechanism separately. 
Specifically, for each semantic hierarchy, \ie, holistic-sentence-level ($RV$), keyword-level ($KWV$), and scene-attribute-level ($BV$), we concatenate a learnable query embedding $\mathbf{p}_{\alpha}^{\beta}$ with the text-derived query $\mathbf{q}_{\alpha}^{\beta}$ before performing cross-attention, where $\beta \in \{RV, ~KWV, ~BV\}$, and $\alpha \in \{T, S\}$ ($T$ denotes temporal branch and $S$ denotes spatial branch).

We first conduct feature aggregation for the temporal branch using average pooling along the spatial dimension. For the holistic-sentence semantic-aligned cross-attention module, we use semantic embeddings of the whole textual reference extracted from the textual encoder of the selected vision-language model as the query tokens and query the holistic-sentence retrieved visual tokens, \ie, $\hat{\mathbf{t}}_V$. 
The query tokens $\mathbf{q}^{RV}_T$ are obtained by applying a linear projection on the previously harvested $\hat{\mathbf{t}}_{R}$, additionally, we concatenate learnable queries $\mathbf{p}^{RV}_T$ with the query tokens to introduce more flexibility of the query and improve the generalizability during training. The key and value tokens, \ie, $\mathbf{k}^{RV}_T$ and $\mathbf{v}^{RV}_T$ are acquired through linear projections on $\hat{\mathbf{t}}_V$. The whole procedure is described in Eq.~\ref{eq:rv}.
\begin{equation}
\label{eq:rv}
    \tilde{\mathbf{z}}^{RV}_T = \mathrm{CA}\big(\mathrm{Concat}[\mathbf{q}^{RV}_T,\,\mathbf{p}^{RV}_T], \mathbf{k}^{RV}_T, \mathbf{v}^{RV}_T\big),
\end{equation}
where CA denotes the cross-attention operation and $\tilde{\mathbf{z}}^{RV}_T$ denotes the aggregated temporal-branch embeddings for holistic-sentence-semantic hierarchy. 

Then we conduct a similar procedure as Eq.~\ref{eq:kvw} using linear projection to project the partial-keyword semantic-retrieved visual token $\mathbf{t}_{KW}$ into $\mathbf{q}_T^{KWV}$ as the query, while key and value tokens, \ie, $\mathbf{k}^{KWV}_T$ and $\mathbf{v}^{KWV}_T$, are harvested by linear projections on the holistic-sentence semantic-retrieved visual tokens $\hat{\mathbf{t}}_V$.
\begin{equation}
\label{eq:kvw}
    \tilde{\mathbf{z}}^{KWV}_T = \mathrm{CA}\big(\mathrm{Concat}[\mathbf{q}^{KWV}_T,\,\mathbf{p}^{KWV}_T],\mathbf{k}^{KWV}_T, \mathbf{v}^{KWV}_T\big),
\end{equation}
where $\mathbf{p}^{KWV}_T$ denotes learnable query prompts for partial-keyword hierarchy.

The scene-attribute-hierarchy semantic-aligned cross-attention is achieved by Eq.~\ref{eq:BV} following the same process procedure according to the previous two semantic hierarchies.
\begin{equation}
\label{eq:BV}
        \tilde{\mathbf{z}}^{BV}_T= \mathrm{CA}
\big(\mathrm{Concat}[\mathbf{q}^{BV}_T,\,\mathbf{p}^{BV}_T], \mathbf{k}^{BV}_T, \mathbf{v}^{BV}_T\big).
\end{equation}
Their average $\mathbf{z}_T$ is taken as in Eq.~\ref{eq:19}, where AVG indicates the mean averaging operation.
\begin{equation}
\label{eq:19}
    \mathbf{z}_T = \text{AVG}\Big(
    \tilde{\mathbf{z}}^{RV}_T + 
    \tilde{\mathbf{z}}^{KWV}_T +
    \tilde{\mathbf{z}}^{BV}_T \Big).
\end{equation}
Analogously, we obtain the spatial-branch representations using Eq.~\ref{eq:20}, where $\tilde{\mathbf{z}}^{RV}_S$, $\tilde{\mathbf{z}}^{KWV}_S$, and $\tilde{\mathbf{z}}^{BV}_S$ are computed following the same aggregation procedure as their temporal counterparts.
\begin{equation}
\label{eq:20}
    \mathbf{z}_S = \text{AVG}\Big(
    \tilde{\mathbf{z}}^{RV}_S + 
    \tilde{\mathbf{z}}^{KWV}_S +
    \tilde{\mathbf{z}}^{BV}_S \Big),
\end{equation}
where $\tilde{\mathbf{z}}^{RV}_S$, $\tilde{\mathbf{z}}^{KWV}_S$, and $\tilde{\mathbf{z}}^{BV}_S$ denote the aggregated spatial tokens for the three aforementioned semantic hierarchies.

The temporal representation $\mathbf{z}_T$ is forwarded to a bounding box regression head $\mathbf{f}^{\text{reg}}_T(\cdot)$ and an action recognition head $\mathbf{f}^{\text{cls}}_T(\cdot)$, according to Eq.~\ref{eq:21}.
\begin{equation}
\label{eq:21}
    \hat{\mathbf{b}}_T = \mathbf{f}^{\text{reg}}_T(\mathbf{z}_T), \quad
    \hat{\mathbf{y}}_T = \mathbf{f}^{\text{cls}}_T(\mathbf{z}_T).
\end{equation}
 While the spatial representation $\mathbf{z}_S$ is passed to their spatial counterparts, according to Eq.~\ref{eq:22}.
\begin{equation}
\label{eq:22}
    \hat{\mathbf{b}}_S = \mathbf{f}^{\text{reg}}_S(\mathbf{z}_S), \quad
    \hat{\mathbf{y}}_S = \mathbf{f}^{\text{cls}}_S(\mathbf{z}_S),
\end{equation}
where $\mathbf{f}^{\text{reg}}_S(\cdot)$ and $\mathbf{f}^{\text{cls}}_S(\cdot)$ denote another regression head and another classification head, respectively.

Finally, the predictions from temporal and spatial branches are fused by averaging according to Eq.~\ref{eq:23}.
\begin{equation}
\label{eq:23}
    \hat{\mathbf{b}} = \text{AVG}\big(\hat{\mathbf{b}}_T + \hat{\mathbf{b}}_S\big), \quad
    \hat{\mathbf{y}} = \text{AVG}\big(\hat{\mathbf{y}}_T + \hat{\mathbf{y}}_S\big),
\end{equation}
where $\hat{\mathbf{b}}$ and $\hat{\mathbf{y}}$ denote the final bounding box and atomic action predictions of the referred person, respectively. Both bounding box regression and action classification inherently require spatial and temporal reasoning, since accurate localization depends on spatial precision as well as temporal consistency, while robust action recognition relies on temporal dynamics contextualized by spatial layouts. By aggregating spatial and temporal features separately and then combining them, the model preserves the distinct contributions of each dimension while enabling richer cross-dimensional cues for both regression and classification.

\subsection{Loss Functions}
\label{sec:loss}
We employ Binary Cross-Entropy (BCE) loss for multi-label classification supervision and Mean Squared Error (MSE) loss for bounding box regression supervision, consistent with all baseline approaches. The BCE loss is formulated as in Eq.~\ref{eq:BCE},
\begin{equation}
\label{eq:BCE}
L_{BCE}(\mathbf{y}, \hat{\mathbf{y}}) = -\frac{1}{N_c} \sum_{i=1}^{N_c} \left[ \mathbf{y}_i \log(\hat{\mathbf{y}}_i) + (1 - \mathbf{y}_i) \log(1 - \hat{\mathbf{y}}_i) \right],
\end{equation}
where $\mathbf{y}$ denotes the multi-hot ground truth vector, $\hat{\mathbf{y}}$ represents the predicted probabilities, and $N_c$ is the number of action categories.

For bounding box regression, the MSE loss is defined in Eq.~\ref{eq:MSE},
\begin{equation}
\label{eq:MSE}
\mathcal{L}_{\mathrm{MSE}}(\mathbf{b},\hat{\mathbf{b}})
= \frac{1}{4}\sum_{j=1}^{4}(\mathbf{b}_j-\hat{\mathbf{b}}_j)^2.
\end{equation}
where $\mathbf{b}$ represents the ground-truth coordinates of the bounding box (top-left and bottom-right corners), $\hat{\mathbf{b}}$ denotes the predicted coordinates, and $j$ indexes the four bounding box parameters.

\section{Experiments}
\subsection{Implementation Details}
\label{sec:impl}
All experiments are conducted on four NVIDIA A100 GPUs. We employ the BertAdam~\cite{kingma2014adam} optimizer with a learning rate $lr = 1e^{-4}$, batch size of $128$, learning rate decay of $0.9$, and a warmup ratio of $0.1$. The model is trained for $40$ epochs on our dataset. During training, the text encoder, visual encoder, object detector, and Q-Former from BLIPv2~\cite{li2023blip} are kept frozen. 
For each level of Mamba-based semantic-visual aggregation, we first adopt a linear projection layer to project the feature dimension from $768$ to $256$ and then apply a single Mamba layer to achieve the state-space modeling. The trainable parameters of the proposed method with the InternVideo2.5 backbone (frozen) are $9.6M$.
For evaluation, we adopt multi-label mean Average Precision (mAP), Area Under the Receiver Operating Characteristic (AUROC) curve , and mean Intersection over Union (mIOU) as evaluation metrics. 

\begin{table*}[t!]
\centering
\caption{Experimental results on both the \texttt{RefAVA} dataset and the \texttt{RefAVA++} dataset. 
Approaches are evaluated using mIOU for referred person localization, mAP and AUROC for referred person atomic action recognition. Performances of the validation set (Val) and test set (Test) on these two datasets are reported.}
\vskip-2ex
\label{tab:ravar_benchmark}
\resizebox{1\textwidth}{!}{
\begin{tabular}{r|c|ccc|ccc|ccc|ccc}
\toprule
\multicolumn{2}{c|}{\multirow{1}{*}{\textbf{Dataset}}}  & \multicolumn{6}{c|}{\multirow{1}{*}{\textbf{RefAVA}}} & \multicolumn{6}{c}{\multirow{1}{*}{\textbf{RefAVA++}}} \\ 
\midrule
\multicolumn{2}{c|}{\multirow{2}{*}{\textbf{Method}}} & \textbf{mIOU} & \textbf{mAP} & \textbf{AUROC} & \textbf{mIOU} & \textbf{mAP} & \textbf{AUROC}  & \textbf{mIOU} & \textbf{mAP} & \textbf{AUROC} & \textbf{mIOU} & \textbf{mAP} & \textbf{AUROC}  \\ 
\cmidrule{3-14}
\multicolumn{2}{c|}{} & \multicolumn{3}{c|}{Val}           & \multicolumn{3}{c|}{Test}  & \multicolumn{3}{c|}{Val}           & \multicolumn{3}{c}{Test} \\ 
\cmidrule{1-14} 
\multirow{5}{*}{AAL}& I3D~\cite{carreira2017quo}  & 0.00 &44.04&57.77&0.00& 44.64 & 62.71  & 0.00& 44.13& 58.79& 0.00&45.26&60.03\\
&X3D~\cite{feichtenhofer2020x3d} &0.26 & 44.45& 59.09& 0.27&46.34&64.51 & 8.91& 44.16 &58.74&9.03&44.57&59.87\\
&MViTv2-B~\cite{li2022mvitv2} & 0.76 &42.32 &56.43 &0.66 &42.60 &59.30  &0.07&48.16&64.21&0.11&48.16&63.66\\
&VideoMAE2-B~\cite{wang2023videomae}& 0.16& 42.02& 55.12& 0.29& 41.87& 59.10 &0.14&43.46&58.73&0.09&43.76&58.76\\
&Hiera-B~\cite{ryali2023hiera} & 0.49& 42.74& 56.72& 0.62& 41.14 &58.70& 0.52& 45.45& 60.90& 0.50& 46.11& 60.90\\
\cmidrule{1-14}
\multirow{2}{*}{VQA} &Singularity~\cite{lei2022revealing} &26.34&42.43 & 59.52 & 29.78&42.18& 56.12  & 27.26&41.14&55.60& 25.30& 42.76& 57.50\\
&AskAnything13B~\cite{2023videochat}  & 20.09 &  51.42  & 66.12  & 22.35  &  52.25 & 69.35 & 27.43& 42.36& 56.98& 26.30& 44.60& 58.96\\
&Qwen2.5-VL-Instruct~\cite{ahmed2025qwen}  & 24.06 & 54.24  & 69.17  & 25.42 & 55.60 & 72.69 & 23.92 & 53.38  & 68.07& 23.94&  52.12& 69.14\\
\cmidrule{1-14}
\multirow{4}{*}{VTR} &MeVTR~\cite{zhang2023multi}  & 30.78 & 38.42 & 51.01 & 29.79&  36.27&  52.45& 29.02&38.24&52.62&29.15&38.24&51.52\\ 
&CLIP4CLIP~\cite{luo2022clip4clip}  & 34.75&39.48 &52.57 &32.33&37.17&55.05& 35.07&40.22&54.03&34.20&40.37&53.89\\
&XClIP~\cite{ma2022x}  &35.54& 42.46 & 56.30 & 31.79 & 40.82& 58.71& 36.26& 51.37&67.15& 35.80& 51.37&67.15\\
&BLIPv2~\cite{li2023blip}  & 32.99 & 52.13& 66.56& 32.75& 53.19& 69.92 & 23.50& 54.58&70.16&23.40& 54.75& 70.22\\
&InternVideo2.5~\cite{wang2025internvideo2}& 38.09 & 58.87 & 73.41 & 39.15& 58.96 &75.83  &32.59& 59.48  &73.95 & 33.53 & 59.09& 74.28\\
\midrule
\multirow{3}{*}{SF} 
&CLIP~\cite{radford2021learning}+SAM~\cite{kirillov2023segment} &  32.95 & 49.74 & 64.55 & 29.80 & 51.34 & 69.11 & 22.34& 55.77&70.51& 22.41& 53.82&69.41\\
&CLIP~\cite{radford2021learning}+DETR~\cite{carion2020end}& 33.96 & 47.57 & 61.83 & 33.84 & 50.92 & 67.29 & 28.80& 51.52&66.70&27.09&48.65&64.15\\
&CLIP~\cite{radford2021learning}+REFCLIP~\cite{jin2023refclip} & 34.43 & 47.79 & 62.50 & 32.28 & 49.90 & 67.44 & 29.44& 50.69& 65.72&30.78&51.07&66.33\\
\midrule
\multirow{1}{*}{VOS} 
&Su \textit{et al.}~\cite{su2023sequence}  & 23.71 & 52.17 & 66.67 & 26.02 & 53.20 & 70.19 &28.91 & 52.35&67.49& 28.07&51.47&67.06\\
&InterRVOS~\cite{jin2026interrvos} & 42.49 & 49.43  & 62.76  &39.05  & 48.47 & 65.97 & 41.13& 56.43& 71.68 & 40.53&58.79 & 73.88\\
\cmidrule{1-14}
\multicolumn{2}{c|}{RefAtomNet  (BLIPv2)~\cite{peng2024referring}} & 38.22 & 55.98 & 69.73 & 36.42 & 57.52 & 73.95& 35.97& 56.54 & 70.97& 35.73&57.01&71.62\\
\multicolumn{2}{c|}{RefAtomNet  (InternVideo2.5)~\cite{peng2024referring}} & 43.63 & 58.11 & 72.18& 42.83 & 58.00 & 72.12 & 41.38& 60.86& 74.88& 39.78& 60.17&73.64\\
\cmidrule{1-14}
\rowcolor{lightgray}\multicolumn{2}{c|}{RefAtomNet++ (BLIPv2)} & 43.71 & 56.83 & 71.27 & 42.52 & \textbf{59.81} & 75.72 & 39.12 & 58.24 & 72.58& 38.58& 58.84& 73.28\\
\rowcolor{lightgray}\multicolumn{2}{c|}{RefAtomNet++ (InternVideo2.5)} & \textbf{50.39}& \textbf{59.21} &  \textbf{73.89} & \textbf{51.14}  & 59.55 & \textbf{76.59} & \textbf{49.49} & \textbf{62.33} & \textbf{76.06} & \textbf{50.33} & \textbf{61.90} & \textbf{76.10}\\
\bottomrule
\end{tabular}
}
\vskip-2ex
\end{table*}

\begin{table}[t!]
\caption{Experimental results for module ablation of \texttt{RefAtomNet++} on the \texttt{RefAVA} dataset.}
\vskip-1ex
\label{tab:module_ablation}
\scalebox{0.95}{\begin{tabular}{l|ccc|ccc}

\toprule
\multirow{2}{*}{\textbf{Method}} & \textbf{mIOU}   & \textbf{mAP}& \textbf{AUROC}&\textbf{mIOU}& \textbf{mAP} & \textbf{AUROC} \\
\cmidrule{2-7}
& \multicolumn{3}{c}{Val} & \multicolumn{3}{|c}{Test}\\
\midrule
BLIPv2~\cite{li2023blip} & 32.99 & 52.13& 66.56& 32.75& 53.19& 69.92 \\
\midrule
w/o TSR-QB  & 36.93 & 55.41& 69.63& 36.93 & 59.47 &74.79\\
w/o BSR-QB  & 38.97 & 55.72& 69.84& 38.60& 58.08& 73.76\\

w/o T-QB & 27.14 & 54.82 &69.14 & 29.00 & 56.83& 73.29\\
\midrule
w/o MHS-CA&39.19  & 55.22 & 69.43& 36.84 & 56.53& 73.97\\
w/o QPrompts & 43.33 & 55.52 & 69.72& 41.40 & 58.50 & 75.13\\
\midrule
\rowcolor{lightgray}Ours & \textbf{43.71} & \textbf{56.83} & \textbf{71.27} & \textbf{42.52} & \textbf{59.81} & \textbf{75.72}\\
\bottomrule
\end{tabular}}
\end{table}

\begin{table}[t!]
\caption{Experimental results for module ablation of \texttt{RefAtomNet++} (InternVideo2.5 backbone~\cite{wang2025internvideo2}) on the \texttt{RefAVA++} dataset.}
\vskip-1ex
\label{tab:module_ablation_internvideo}
\scalebox{0.9}{\begin{tabular}{l|ccc|ccc}

\toprule
\multirow{2}{*}{\textbf{Method}} & \textbf{mIOU}   & \textbf{mAP}& \textbf{AUROC}&\textbf{mIOU}& \textbf{mAP} & \textbf{AUROC} \\
\cmidrule{2-7}
& \multicolumn{3}{c}{Val} & \multicolumn{3}{|c}{Test}\\
\midrule
InternVideo2.5~\cite{wang2025internvideo2}&32.59& 59.48  &73.95 & 33.53 & 59.09& 74.28\\
\midrule
w/o TSR-QB  & 34.96 & 59.35& 73.99& 35.98& 59.04& 74.25\\
w/o BSR-QB  & 37.83 &60.10 & 74.66& 39.11& 59.66& 74.79\\

w/o T-QB  & 42.11 & 60.90& 75.27& 43.53& 60.37&75.35 \\
\midrule
w/o MHS-CA & 39.47 & 60.53 & 75.00 & 40.78 & 60.08 & 75.00\\
w/o QPrompts & 42.36 & 61.25 &75.54  & 43.72 & 60.58 & 75.56\\
\midrule
\rowcolor{lightgray}Ours & \textbf{49.49} & \textbf{62.33} & \textbf{76.06} & \textbf{50.33} & \textbf{61.90} & \textbf{76.10}\\
\bottomrule
\end{tabular}}
\end{table}

\begin{table}[t!]
\caption{Ablation of temporal and spatial reasoning branches of \texttt{RefAtomNet++} on the \texttt{RefAVA} dataset.}
\vskip-2ex
\label{tab:temporal_spatial_abl}
\scalebox{0.95}{\begin{tabular}{l|ccc|ccc}

\toprule
\multirow{2}{*}{\textbf{Method}} & \textbf{mIOU}   & \textbf{mAP}& \textbf{AUROC}&\textbf{mIOU}& \textbf{mAP} & \textbf{AUROC} \\
\cmidrule{2-7}
& \multicolumn{3}{c}{Val} & \multicolumn{3}{|c}{Test}\\
\midrule
BLIPv2~\cite{li2023blip} & 32.99 & 52.13& 66.56& 32.75& 53.19& 69.92 \\
\midrule
w/o TA & 40.51 & 55.71 &70.15 & 39.81 &57.91 & 74.33\\
w/o SA& 40.40 & 55.37 &69.95 &39.31  &57.66 & 74.53\\
\midrule
\rowcolor{lightgray}Ours & \textbf{43.71} & \textbf{56.83} & \textbf{71.27} & \textbf{42.52} & \textbf{59.81} & \textbf{75.72}\\
\bottomrule
\end{tabular}}
\end{table}

\begin{table}[t!]
\caption{Ablation studies on different trajectory aggregation strategies. The experiments are conducted on the \texttt{RefAVA} dataset.}

\label{tab:abl_agg}
\scalebox{0.93}{\begin{tabular}{l|ccc|ccc}

\toprule
\multirow{2}{*}{\textbf{Method}} & \textbf{mIOU}   & \textbf{mAP}& \textbf{AUROC}&\textbf{mIOU}& \textbf{mAP} & \textbf{AUROC} \\
\cmidrule{2-7}
& \multicolumn{3}{c}{Val} & \multicolumn{3}{|c}{Test}\\
\midrule
LSTM  & 38.43 & 55.54& 69.73& 37.05&57.32 &74.45 \\
Linear Projection & 36.00 & 55.51 & 69.89 & 34.83 &58.09 & 74.44\\
Transformer & 39.46&  55.64& 69.98 & 39.06 & 58.35& 74.76\\
\midrule
\rowcolor{lightgray}Ours & \textbf{43.71} & \textbf{56.83} & \textbf{71.27} & \textbf{42.52} & \textbf{59.81} & \textbf{75.72}\\
\bottomrule
\end{tabular}}
\end{table}

\begin{table}[t!]
\caption{Comparison with other multi-modal fusion strategies on the \texttt{RefAVA} dataset.}
\vskip-2ex
\label{tab:mm}
\scalebox{0.85}{\begin{tabular}{l|ccc|ccc}
\toprule
\multirow{2}{*}{\textbf{Fusion}} & \textbf{mIOU}   & \textbf{mAP}& \textbf{AUROC}&\textbf{mIOU}& \textbf{mAP} & \textbf{AUROC} \\
\cmidrule{2-7}
& \multicolumn{3}{c}{Val} & \multicolumn{3}{|c}{Test}\\
\midrule
Addition &27.30 &50.70  &65.31  &29.09  &51.26 &68.20 \\
Concatenation  &18.64 & 52.23 &66.45 & 20.65& 53.44& 70.70\\
Multiplication  & 23.90 & 51.55 & 65.66& 25.05 & 53.33 & 70.48\\
AttentionBottleneck~\cite{nagrani2021attention}  &33.47  & 50.97 &65.07 &33.02 &54.02  &71.08 \\
McOmet~\cite{zong2023mcomet}  &23.88  & 51.58 &65.65 &25.02  & 53.21 &70.42 \\
\midrule
RefAtomNet  & 38.22 & 55.98 & 69.73 & 36.42 & 57.52 & 73.95\\

\rowcolor{lightgray}RefAtomNet++ & \textbf{43.71} & \textbf{56.83} & \textbf{71.27} & \textbf{42.52} & \textbf{59.81} & \textbf{75.72}\\
\bottomrule
\end{tabular}}
\end{table}

\begin{table}[t!]
\caption{Experimental results on test-time rephrasing. The experiments are conducted on the \texttt{RefAVA} dataset.}
\vskip-2ex
\label{tab:rephrase}
\scalebox{0.9}{\begin{tabular}{l|ccc|ccc}
\toprule
\multirow{2}{*}{\textbf{Method}} & \textbf{mIOU}   & \textbf{mAP}& \textbf{AUROC}&\textbf{mIOU}& \textbf{mAP} & \textbf{AUROC} \\
\cmidrule{2-7}
& \multicolumn{3}{c}{Val} & \multicolumn{3}{|c}{Test}\\
\midrule
Singularity~\cite{lei2022revealing}  & 18.45 & 41.27 & 58.47 & 20.54 & 41.39 & 55.32\\
XCLIP~\cite{ma2022x}  & 31.95 & 41.35  & 54.35  & 29.84  & 40.74  & 58.45 \\
AskAnything~\cite{2023videochat}  & 19.74 & 51.11 & 65.83& 21.60 & 51.96 & 69.04\\
BLIPv2~\cite{li2023blip}  & 31.00 & 51.45 & 65.88 & 31.34 & 52.35 & 68.87 \\
\midrule
RefAtomNet  & 34.65 & 55.75 &69.52 & 33.14  & 57.23  &73.76 \\
\rowcolor{lightgray}RefAtomNet++  & \textbf{40.07} & \textbf{57.10} & \textbf{71.54} &  \textbf{39.67} & \textbf{58.82}  & \textbf{75.26}\\
\bottomrule
\end{tabular}}
\end{table}

\begin{table}[t!]
\caption{Generalizability to different visual-textual encoder architectures. The experiments are conducted on the \texttt{RefAVA} dataset.}
\vskip-2ex
\label{tab:gen_tvr}
\scalebox{0.85}{\begin{tabular}{l|ccc|ccc}
\toprule
\multirow{2}{*}{\textbf{Method}} & \textbf{mIOU}   & \textbf{mAP}& \textbf{AUROC}&\textbf{mIOU}& \textbf{mAP} & \textbf{AUROC} \\
\cmidrule{2-7}
& \multicolumn{3}{c}{Val} & \multicolumn{3}{|c}{Test}\\
\midrule
XCLIP~\cite{ma2022x}  &35.54& 42.46 & 56.30 & 31.79 & 40.82& 58.71\\
RefAtomNet (XCLIP) & 38.59 & 47.40 & 61.20& 36.61 & 48.59& 66.47 \\
RefAtomNet++ (XCLIP)  & \textbf{44.12}& \textbf{53.45} &\textbf{67.55} &\textbf{42.24} &\textbf{53.03} & \textbf{70.50}\\
\midrule
BLIPv2~\cite{li2023blip}  & 32.99 & 52.13& 66.56& 32.75& 53.19& 69.92 \\
RefAtomNet (BLIPv2) & 38.22 & 55.98 & 69.73 & 36.42 & 57.52 & 73.95\\
\rowcolor{lightgray}RefAtomNet++ (BLIPv2) & \textbf{43.71} & \textbf{56.83} & \textbf{71.27} & \textbf{42.52}& \textbf{59.81} &\textbf{75.72} \\
\bottomrule
\end{tabular}}
\vskip-3ex
\end{table}

\begin{table}[t]
\caption{Generalizability to other object detectors. The experiments are conducted on the \texttt{RefAVA} dataset.}
\vskip-2ex
\label{tab:detector}
\centering
\scalebox{0.76}{\begin{tabular}{l|ccc|ccc}
\toprule
\multirow{2}{*}{\textbf{Method}} & \textbf{mIOU}   & \textbf{mAP}& \textbf{AUROC}&\textbf{mIOU}& \textbf{mAP} & \textbf{AUROC} \\
\cmidrule{2-7}
& \multicolumn{3}{c}{Val} & \multicolumn{3}{|c}{Test}\\
\midrule
BLIPv2~\cite{li2023blip} & 32.99 & 52.13& 66.56& 32.75& 53.19& 69.92 \\
RefAtomNet (RetinaNet~\cite{lin2017focal})  & 38.65 & 55.08 &69.16& 37.82 & 56.67 &73.63 \\
RefAtomNet (DETR~\cite{carion2020end}) & 38.22 & 55.98 & 69.73 & 36.42 & 57.52 & 73.95\\

RefAtomNet++ (RetinaNet~\cite{lin2017focal})  & 41.65 & 56.04 &70.34 & 40.06 & \textbf{59.85} & 75.44\\
\rowcolor{lightgray}RefAtomNet++ (DETR~\cite{carion2020end}) & \textbf{43.71} & \textbf{56.83} & \textbf{71.27} & \textbf{42.52} & 59.81 & \textbf{75.72}\\
\bottomrule
\end{tabular}}
\vskip-2ex
\end{table}

\begin{table}[t!]
\caption{Ablation study on the number of learnable queries conducted on the \texttt{RefAVA} dataset.}
\vskip-2ex
\label{tab:n_q}
\centering
\begin{tabular}{l|ccc|ccc}
\toprule
\multirow{2}{*}{\textbf{\#Queries}} & \textbf{mIOU}   & \textbf{mAP}& \textbf{AUROC}&\textbf{mIOU}& \textbf{mAP} & \textbf{AUROC} \\
\cmidrule{2-7}
& \multicolumn{3}{c}{Val} & \multicolumn{3}{|c}{Test}\\
\midrule
2  & 39.70 & 56.65& 70.71& 41.34 & 59.31 & 75.13\\
4  & 42.75 &  56.69& 71.22 & 42.20 & \textbf{59.83} & \textbf{76.07}\\
\rowcolor{lightgray}6 & \textbf{43.71} & \textbf{56.83} & \textbf{71.27} & \textbf{42.52} & 59.81 & 75.72\\
8  & 42.90 & 56.41 & 70.86 & 41.88 & 59.35 & 75.37\\
10  & 42.31 & 56.55 & 70.95 & 41.30 & 59.51 & 75.37\\
12   & 41.60 & 56.48 & 70.81 & 40.55 & 59.69 & 75.55\\
\bottomrule
\end{tabular}
\vskip-2ex
\end{table}

\begin{table}[t!]
\caption{Ablation of the frame number. The experiments are conducted on the \texttt{RefAVA} dataset.}
\vskip-1ex
\label{tab:n_f}
\centering
\begin{tabular}{l|ccc|ccc}
\toprule
\multirow{2}{*}{\textbf{\#Frames}} & \textbf{mIOU}   & \textbf{mAP}& \textbf{AUROC}&\textbf{mIOU}& \textbf{mAP} & \textbf{AUROC} \\
\cmidrule{2-7}
& \multicolumn{3}{c}{Val} & \multicolumn{3}{|c}{Test}\\
\midrule
2  & 36.02 & 55.87 & 69.97 & 35.11 & 59.33 & 74.76\\
4  & 39.32 & 56.39 & 70.59 & 38.66 & 60.06 & 75.18\\
6  & 42.10 & 56.71& 71.18 & 41.33 & \textbf{60.16} & 75.63\\
\rowcolor{lightgray}8  & \textbf{43.71} & \textbf{56.83} & \textbf{71.27} & \textbf{42.52} & 59.81 & \textbf{75.72}\\
10  & 42.51 & 56.74 & 71.19 & 41.58 & 60.07 & 75.60\\
12  & 42.04 & 56.74 & 71.20 & 41.21 & 60.13 & 75.63\\
\bottomrule
\end{tabular}
\vskip-2ex
\end{table}

\begin{table}[t!]
\caption{Evaluation of the quality of the extracted semantic attribute trajectories. Hit@1 is the mean rate that the selected top-1 visual token center falls inside the target box, with ci95 (low) and ci95 (high) giving the lower and upper bounds of its 95\% clip-level bootstrap confidence interval.}
\label{tab:traj_eval}
\centering
\begin{tabular}{llll}
\toprule
\textbf{Method}        & \textbf{Hit@1}          & \textbf{Ci95 (low)}       & \textbf{Ci95 (high)}      \\
\midrule
Random        & 37.50          & 37.20          & 37.79          \\
\textbf{Ours} & \textbf{55.81} & \textbf{54.63} & \textbf{56.90} \\
\bottomrule
\end{tabular}
\vskip-3ex
\end{table}

\begin{figure*}
    \centering
    \includegraphics[width=1\linewidth]{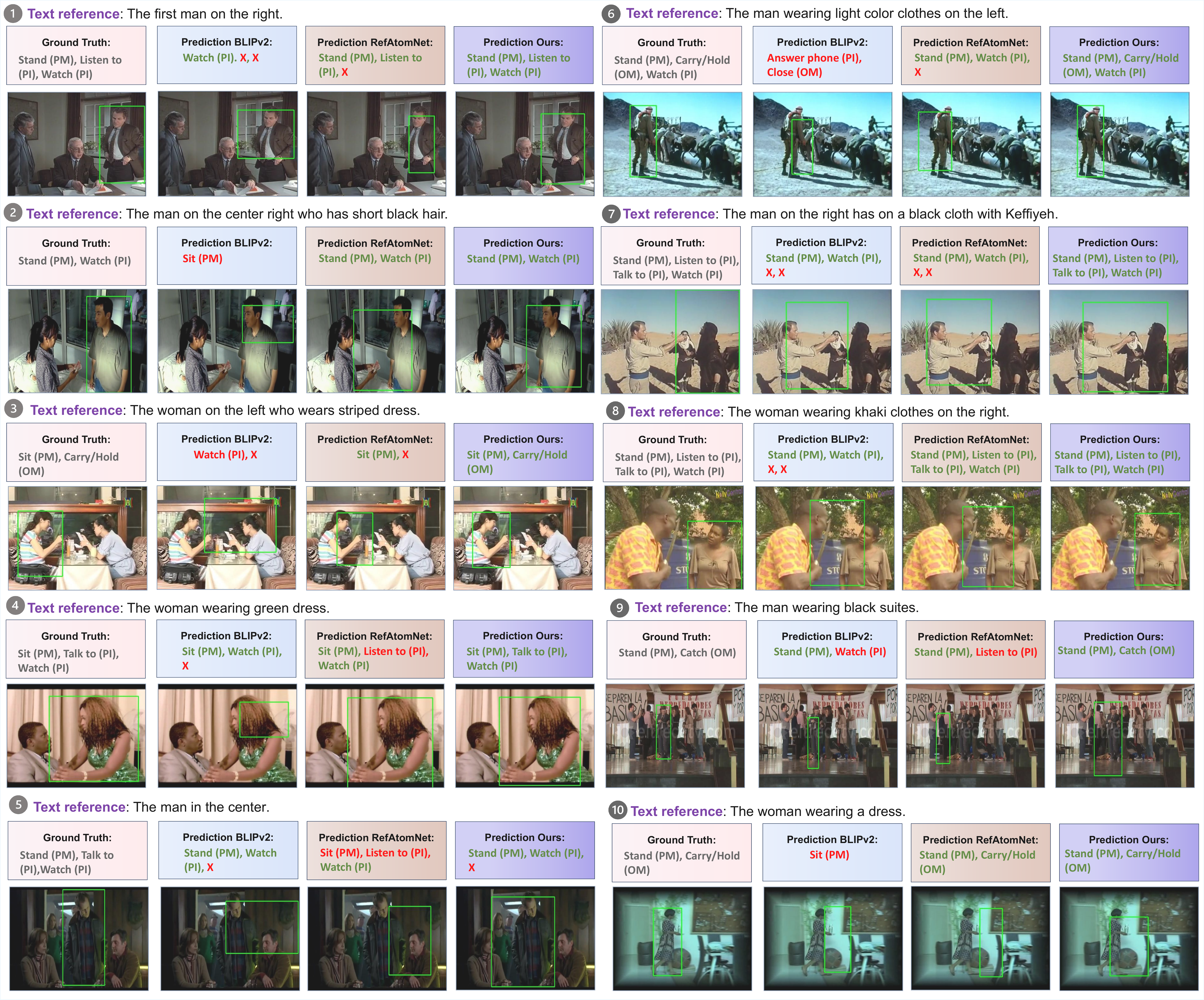}
    \vskip-2ex
    \caption{Qualitative comparisons of BLIPv2~\cite{li2023blip},
    \texttt{RefAtomNet}, and \texttt{RefAtomNet++} on the \texttt{RefAVA} test
    set using BLIPv2 as the common backbone. Each example displays the
    textual reference, ground-truth actions, predicted actions, and localized
    subject. \texttt{RefAtomNet++} recovers the complete action set in nine
    of the ten examples. Example 5 presents a remaining failure: the model
    correctly grounds the referred person but misses the relational action
    \emph{Talk to}, showing that accurate grounding does not necessarily
    imply complete multi-label action recognition.}
    \vskip-2ex
    \label{fig:qualitative}
\end{figure*}

\subsection{Analyses of the Benchmark Results}
\label{sec:benchmark}
Table~\ref{tab:ravar_benchmark} compares \texttt{RefAtomNet++} with representative methods from Atomic Action Localization (AAL), Video Question Answering (VQA), Video-Text Retrieval (VTR), Single-Frame (SF) modeling, Video Object Segmentation (VOS), and our previous \texttt{RefAtomNet}~\cite{peng2024referring}. All paired results below are reported in validation/test order. Across both datasets, AAL methods obtain non-trivial action-recognition scores but substantially lower mIoU, showing that action-centric visual features alone cannot identify a language-referred person. Vision--language models from VQA and VTR improve this alignment, whereas SF and VOS methods exhibit complementary strengths in either recognition or localization but do not fully address person-specific spatio-temporal reasoning.

\noindent\textbf{RefAVA.}
Under the controlled BLIPv2 setting and a comparable parameter budget, \texttt{RefAtomNet++} improves over \texttt{RefAtomNet} by $5.49\%/6.10\%$ in mIoU, $0.85\%/2.29\%$ in mAP, and $1.54\%/1.77\%$ in AUROC. These gains primarily stem from the proposed multi-level semantic trajectory retrieval and fusion, rather than increased model capacity. With the InternVideo2.5 backbone, \texttt{RefAtomNet++}
achieves $50.39\%/51.14\%$ mIoU, $59.21\%/59.55\%$ mAP, and
$73.89\%/76.59\%$ AUROC on the \texttt{RefAVA} validation/test sets.

\noindent\textbf{RefAVA++.}
The controlled \texttt{BLIPv2} backbone comparison shows the same trend: \texttt{RefAtomNet++} surpasses \texttt{RefAtomNet} by $3.15\%/2.85\%$ in mIoU, $1.70\%/1.83\%$ in mAP, and $1.61\%/1.66\%$ in AUROC. When equipped with \texttt{InternVideo2.5}, it establishes the best results in Table~\ref{tab:ravar_benchmark}, achieving $49.49\%/50.33\%$ mIoU, $62.33\%/61.90\%$ mAP, and $76.06\%/76.10\%$ AUROC. Relative to the standalone InternVideo2.5 backbone equipped with RAVAR prediction head, the corresponding gains are $16.90\%/16.80\%$, $2.85\%/2.81\%$, and $2.11\%/1.82\%$, confirming that the framework remains effective on the larger and more challenging benchmark.

Overall, the substantially larger improvements in mIoU suggest that partial-keyword, scene-attribute, and holistic-sentence retrieval primarily strengthen target disambiguation and suppress irrelevant visual evidence. The consistent mAP and AUROC gains are further in line with the intended roles of Mamba-based state-space aggregation and the separate spatial and temporal reasoning branches, which preserve long-range action dynamics after the relevant person-specific trajectories have been selected. The results therefore demonstrate both architecture-level improvements under controlled backbones and effective scalability to stronger video-language representations.

\subsection{Module Ablations}
\label{sec:abl}
\noindent\textbf{Analysis of each component.}
Table~\ref{tab:module_ablation} isolates the roles of the three semantic hierarchies and their reintegration in \texttt{RefAtomNet++}. The \textit{w/o TSR-QB}, \textit{w/o BSR-QB}, and \textit{w/o T-QB} variants remove the partial-keyword, scene-attribute, and holistic-sentence pathways, respectively, together with their trajectory aggregation and semantic-aligned cross-attention. The \textit{w/o MHS-CA} variant retains the hierarchical trajectories but removes their reintegration with the dense video tokens, whereas \textit{w/o QPrompts} removes the learnable prompts from all cross-attention pathways.

Removing holistic-sentence reasoning produces the largest localization loss: validation/test mIoU decreases from $43.71\%/42.52\%$ to $27.14\%/29.00\%$, whereas mAP decreases to $54.82\%/56.83\%$. Thus, keywords and scene attributes cannot replace the compositional constraint encoded by the complete expression. For example, ``the first man on the right'' requires the entity, ordinal, and spatial relation to be bound to the same person.
The corresponding case in Fig.~\ref{fig:qualitative} illustrates that this global constraint is particularly important in multi-person scenes.

The other two hierarchies have more specialized roles. Removing partial-keyword reasoning decreases validation/test mIoU by $6.78\%/5.59\%$ but test mAP by only $0.34\%$. Appearance and position words therefore primarily distinguish the referent from nearby people, especially when candidates have similar global characteristics. In contrast, removing scene-attribute reasoning decreases mIoU by $4.74\%/3.92\%$ and test mAP by $1.73\%$. Its smaller localization loss but larger recognition loss indicates that detected objects and their spatial arrangement provide contextual evidence for person--object actions. This interpretation is consistent with the qualitative example 3 in Figure~\ref{fig:qualitative}, where \texttt{RefAtomNet++} recovers \textit{Carry/Hold}, which \texttt{RefAtomNet} misses. A likely failure mode is nevertheless imposed by the center-frame detector: small, occluded, or undetected objects cannot contribute to the scene trajectories.

The \textit{w/o MHS-CA} results further show that trajectory construction alone is insufficient. Removing MHS-CA decreases validation/test mIoU by $4.52\%/5.68\%$, mAP by $1.61\%/3.28\%$, and AUROC by $1.84\%/1.75\%$. The compact trajectories select cue-relevant evidence but may discard spatial details required by the prediction heads; MHS-CA recovers these details by using each hierarchical representation to query the original dense video tokens. Removing the query prompts causes a smaller but consistent test degradation of $1.12\%$, $1.31\%$, and $0.59\%$ in mIoU, mAP, and AUROC, respectively. The prompts therefore improve task-adaptive retrieval, but the larger hierarchy-removal losses show that the main gains arise from semantic organization rather than additional parameters.

Table~\ref{tab:module_ablation_internvideo} evaluates the components of \texttt{RefAtomNet++} with the InternVideo2.5 backbone on \texttt{RefAVA++}.
The complete model achieves the best performance across all metrics on both splits, reaching $49.49\%/50.33\%$ mIoU, $62.33\%/61.90\%$ mAP, and $76.06\%/76.10\%$ AUROC on the validation/test sets.

Overall, the results reveal a division of labor: holistic reasoning preserves the complete referential constraint, keyword trajectories resolve competing people, scene trajectories supply contextual action evidence, and MHS-CA reconnects these selected cues with spatially detailed features.

\noindent\textbf{Analysis of ablation experiments regarding the temporal and spatial aggregation branch.} 
Table~\ref{tab:temporal_spatial_abl} reports ablation results of \texttt{RefAtomNet++} on the \texttt{RefAVA} dataset by individually removing the temporal and spatial reasoning branches. Without the temporal aggregation branch (\textit{w/o TA}), performance drops from $42.52\%$ to $39.81\%$ mIoU and from $59.81\%$ to $57.91\%$ mAP on the test set, indicating that temporal reasoning is critical for capturing motion cues and action continuity. Similarly, removing the spatial aggregation branch (\textit{w/o SA}) reduces mIoU to $39.31\%$ and mAP to $57.66\%$, showing the importance of spatial reasoning for precise localization of the referred individual and the prediction of the atomic-level actions. While both branches contribute significantly, their joint integration yields the best results, with \texttt{RefAtomNet++} achieving $42.52\%$ mIoU, $59.81\%$ mAP, and $75.72\%$ AUROC on the test set, outperforming all the variants.

\noindent\textbf{Analysis of different trajectory aggregation methods.}
Table~\ref{tab:abl_agg} reports ablation results on different strategies for semantic trajectory aggregation. Compared to LSTM, Linear Projection, and Transformer variants, our Mamba-based approach consistently achieves the best results across all metrics and sets. Specifically, it improves validation mIOU by $+4.25\%$ over the best baseline (Transformer) and test mIOU by $+3.46\%$, showing stronger spatial grounding ability. On action recognition, Mamba also outperforms alternatives, reaching $56.83\%$/$59.81\%$ mAP on val/test, surpassing Transformer’s $55.64\%$/$58.35\%$. AUROC follows the same trend, with Mamba achieving $71.27\%$/$75.72\%$, again higher than all baselines. These results highlight that the selective state-space modeling of Mamba better captures long-range temporal dependencies and integrates semantic cues, making it more effective for fine-grained RAVAR. State-space modeling offers a continuous and memory-efficient way to aggregate semantic trajectories by capturing long-range temporal dependencies with stable dynamics. Unlike the other ablated counterparts, it enables smooth temporal filtering and considers coherent visual-semantic cues from different hierarchies that ensure a better alignment between the given textual reference and motion cues of the desired person.

\noindent\textbf{Comparison with other multi-modal fusion strategies.}
Table~\ref{tab:mm} presents a comparison of \texttt{RefAtomNet++} against alternative multi-modal fusion strategies on the \texttt{RefAVA} dataset. Simple fusion methods such as addition, concatenation, and multiplication perform poorly, with validation and test mIoU scores below $27.30\%$ and $29.09\%$, respectively, reflecting their inability to capture complex cross-modal dependencies. More sophisticated approaches, including AttentionBottleneck~\cite{nagrani2021attention} and McOmet~\cite{zong2023mcomet}, achieve better results but still lag behind \texttt{RefAtomNet}~\cite{peng2024referring}, with $36.42\%$ mIoU, $57.52\%$ mAP, and $73.95\%$ AUROC on the test set. 
In contrast, \texttt{RefAtomNet++} with BLIPv2 backbone~\cite{li2023blip} achieves better performance with $42.52\%$ mIoU, $59.81\%$ mAP, and $75.72\%$ AUROC on the test set, surpassing all fusion baselines by a substantial margin. 
This improvement stems from its multi-hierarchical semantic-aligned cross-attention and multi-trajectory Mamba-based aggregation, which enable more effective fine-grained alignment of textual and visual cues across spatial and temporal dimensions from holistic-sentence, partial-keyword, and scene-attribute perspectives.

\noindent\textbf{Analysis of the generalizability towards test-time rephrasing.}
Table~\ref{tab:rephrase} presents the experimental results under the challenging setting of test-time rephrasing, where each textual query is rephrased three times using ChatGPT~\cite{openai2022chatgpt} and the averaged performance is reported. This setting evaluates model robustness to linguistic variations, a critical aspect for practical deployment in natural human-computer interaction scenarios. 
Existing multimodal baselines, including Singularity~\cite{lei2022revealing}, XCLIP~\cite{ma2022x}, AskAnything~\cite{2023videochat}, and BLIPv2~\cite{li2023blip}, show considerable sensitivity to rephrasing, with mIoU values dropping below $31.34\%$ on the test set. \texttt{RefAtomNet} demonstrates improved robustness, maintaining $33.14\%$ mIoU, $57.23\%$ mAP, and $73.76\%$ AUROC despite the linguistic shifts. \texttt{RefAtomNet++} achieves the best overall results, boosting performance to $39.67\%$ mIoU, $58.82\%$ mAP, and $75.26\%$ AUROC on the test set. 
These results highlight the effectiveness of our multi-trajectory semantic-retrieval Mamba and multi-hierarchical semantic-aligned cross-attention design in mitigating the adverse effects of query reformulation, thereby enabling more reliable RAVAR with natural language variability.

\noindent\textbf{Analysis of the generalizability to different visual-textual encoders.} 
Table~\ref{tab:gen_tvr} presents the evaluation of \texttt{RefAtomNet++} with different visual-textual encoder architectures to assess its generalizability. 
Using both XCLIP~\cite{ma2022x} and BLIPv2~\cite{li2023blip}, \texttt{RefAtomNet++} consistently outperforms the corresponding \texttt{RefAtomNet} baselines. 
For instance, with the XCLIP encoder, \texttt{RefAtomNet++} improves mIoU from $36.61\%$ to $42.24\%$, mAP from $48.59\%$ to $53.03\%$, and AUROC from $66.47\%$ to $70.50\%$ on the test set. Similarly, with BLIPv2, it also achieves significant gains, raising mIoU from $36.42\%$ to $42.52\%$, mAP from $57.52\%$ to $59.81\%$, and AUROC from $73.95\%$ to $75.72\%$. \texttt{RefAtomNet++} generalizes well across different vision-language backbones because its multi-trajectory semantic-retrieval and multi-hierarchical cross-attention modules provide a flexible fusion mechanism from different semantic hierarchies that complements and enhances diverse encoder representations.

\noindent\textbf{Analysis of the generalizability towards different object detectors.}
Table~\ref{tab:detector} evaluates the generalizability of \texttt{RefAtomNet++} across different object detectors, comparing RetinaNet~\cite{lin2017focal} and DETR~\cite{carion2020end}. 
As shown, both \texttt{RefAtomNet} and \texttt{RefAtomNet++} outperform the BLIPv2~\cite{li2023blip} baseline regardless of the underlying detector, indicating that our framework is not tied to a specific detection backbone. With RetinaNet, \texttt{RefAtomNet++} achieves $40.06\%$ mIoU, $59.85\%$ mAP, and $75.44\%$ AUROC on the test set, already surpassing \texttt{RefAtomNet} with the same detector. When \texttt{RefAtomNet++} is coupled with DETR instead of RetinaNet, the test mIoU and AUROC improve by $2.46\%$ and $0.28\%$, respectively, while the test mAP decreases slightly by $0.04\%$. These findings demonstrate that the proposed semantic-retrieval Mamba and multi-hierarchical semantic-aligned cross-attention design of \texttt{RefAtomNet++} remains effective across different detection paradigms, confirming its robustness and adaptability in diverse visual grounding pipelines.

\noindent\textbf{Analysis of the ablation for learnable queries.}
Table~\ref{tab:n_q} reports the ablation study on the number of learnable queries used in \texttt{RefAtomNet++}. We vary the query number from $2$ to $12$. 
Using too few queries (\textit{e.g.}, $2$) leads to insufficient learnable flexibility, while too many queries (\textit{e.g.}, $10$ or $12$) result in redundancy and diluted attention, both causing degraded performance. The best trade-off is achieved with $6$ queries, where \texttt{RefAtomNet++} obtains the highest overall scores of $43.71\%$ mIoU, $56.83\%$ mAP, and $71.27\%$ AUROC on the validation set and $42.52\%$ mIoU, $59.81\%$ mAP, and $75.72\%$ AUROC on the test set. These results suggest that an intermediate query number offers a balanced capacity to capture fine-grained action correspondences while avoiding over-fragmentation of attention.

\noindent\textbf{Analysis of the ablation for the number of temporal frames.}
Table~\ref{tab:n_f} presents the ablation study for the number of frames sampled per video on \texttt{RefAtomNet++}. 
We experiment with frame counts ranging from $2$ to $12$ and observe that both very short clips and overly long clips lead to suboptimal results. Using only $2$ or $4$ frames fails to capture sufficient temporal context, while extending to $10$ or $12$ frames introduces redundancy and noise, slightly degrading performance. 
The optimal balance is achieved with $8$ frames, where the model reaches the best trade-off between spatiotemporal coverage and efficiency.
This demonstrates that moderate temporal sampling provides enough motion cues for fine-grained action reasoning without overwhelming the cross-modal alignment process.

\noindent\textbf{Quality of semantic trajectory retrieval.} On the \texttt{RefAVA++} validation set, we evaluate whether the visual token retrieved by each person-referential keyword falls inside the target-person box as shown in Table~\ref{tab:traj_eval}. \texttt{RefAtomNet++} achieves a Hit@1 of $55.81\%$, outperforming uniform random selection by $18.31\%$. The non-overlapping clip-level bootstrap confidence intervals confirm that the improvement is statistically reliable. Because \texttt{RefAVA++} provides center-frame spatial annotations, this experiment establishes above-chance keyframe grounding rather than full temporal identity consistency.

\subsection{Analyses of the Qualitative Results}
\label{sec:qualitative}
To separate person-grounding errors from action-recognition errors, we analyze predictions in two stages. A predicted box with intersection-over-union below $0.5$ is treated as a grounding failure; for correctly grounded cases, we separately inspect false-positive and false-negative action labels. 
This distinction clarifies whether an action error arises from selecting the wrong person or misinterpreting the behavior of the correct person. We group the dominant errors into four factors: ambiguous references, occlusion or visual similarity, weak object or interaction evidence, and insufficient temporal evidence.

Fig.~\ref{fig:qualitative} compares BLIPv2~\cite{li2023blip}, \texttt{RefAtomNet}, and \texttt{RefAtomNet++} using the same BLIPv2 backbone. Across the indoor scenes (examples 1--4 and 10), outdoor interactions (examples 6--8), and the crowded scene (example 9), \texttt{RefAtomNet++} benefits from combining appearance, spatial, and scene cues. 
In example 3, it recovers both \emph{Sit} and \emph{Carry/Hold} for the woman in the striped dress, whereas BLIPv2 uses incorrect visual evidence and \texttt{RefAtomNet} misses the object-manipulation label. Examples 7 and 8 further require the concurrent recognition of \emph{Stand}, \emph{Listen to}, \emph{Talk to}, and \emph{Watch}; recovering the complete action sets indicates that cue-conditioned trajectories preserve complementary evidence that sentence-level guidance may overlook. 
Overall, \texttt{RefAtomNet++} recovers the complete ground-truth action set in nine of the ten examples.

Example 5 exposes a remaining failure. The model correctly grounds ``the man in the center'' and recognizes \emph{Stand} and \emph{Watch}, but misses \emph{Talk to}. Thus, correct localization does not guarantee complete action recognition. Unlike standing or watching, talking depends on subtle relational evidence that may appear only briefly. Because action terms are excluded from the reference, this interaction must be inferred entirely from the sampled frames. The model therefore remains less reliable for short-duration or relational actions despite retaining the target identity.

The architecture suggests several additional failure conditions. Partial-keyword retrieval selects the nearest spatial token for each cue at every time step; when people share similar clothing or positions, this hard selection may switch to a distractor during occlusion or rapid motion. Mamba can suppress isolated errors but cannot recover identity when incorrect retrieval persists. Moreover, the scene branch uses objects detected in the center frame, so small or occluded objects may be absent, weakening object-dependent actions such as \emph{Answer phone}, \emph{Carry/Hold}, and \emph{Catch}. Uniform sampling may also miss brief events: Table~\ref{tab:n_f} shows that two or four frames provide insufficient evidence, whereas longer sampling introduces redundancy. Finally, averaging the semantic hierarchies and spatial-temporal branches may attenuate a weak but correct cue when their predictions disagree.

\section{Conclusion}
We introduced \texttt{RefAVA++}, a movie-disjoint RAVAR benchmark with over $2.9$ million frames and $75{,}111$ annotated person instances, and proposed \texttt{RefAtomNet++}. 
The model combines cue-conditioned visual token sequences, Mamba-based temporal aggregation, and multi-hierarchical semantic-aligned cross-attention. 
Experiments on \texttt{RefAVA} and \texttt{RefAVA++} demonstrate improved localization and action recognition, with the highest action mAP among the evaluated methods. 
Keyframe analysis further supports above-chance target grounding, while direct evaluation of temporal identity consistency remains future work.

\ifCLASSOPTIONcompsoc
  \section*{Acknowledgments}
\else
  \section*{Acknowledgment}
\fi
The project is funded by the Deutsche Forschungsgemeinschaft (DFG, German Research Foundation) – SFB 1574 – 471687386. This work is also supported in part by the SmartAge project sponsored by the Carl Zeiss Stiftung (P2019-01-003; 2021-2026), the University of Excellence through the ``KIT Future Fields'' project, in part by the Helmholtz Association Initiative and Networking Fund on the HoreKA@KIT partition and the state of Baden-Württemberg through bwHPC and the German Research Foundation (DFG) through grant INST 35/1597-1 FUGG.
This work was also supported in part by the National Natural Science Foundation of China (Grant No. 62473139 and No. 62388101), in part by the Hunan Provincial Research and Development Project (Grant No. 2025QK3019), and in part by the State Key Laboratory of Autonomous Intelligent Unmanned Systems (the opening project number ZZKF2025-2-10). 
This research was partially funded by the Ministry of Education and Science of Bulgaria (support for INSAIT, part of the Bulgarian National Roadmap for Research Infrastructure).

\bibliographystyle{IEEEtran}
\bibliography{bib}

\begin{thebibliography}{10}
\providecommand{\url}[1]{#1}
\csname url@samestyle\endcsname
\providecommand{\newblock}{\relax}
\providecommand{\bibinfo}[2]{#2}
\providecommand{\BIBentrySTDinterwordspacing}{\spaceskip=0pt\relax}
\providecommand{\BIBentryALTinterwordstretchfactor}{4}
\providecommand{\BIBentryALTinterwordspacing}{\spaceskip=\fontdimen2\font plus
\BIBentryALTinterwordstretchfactor\fontdimen3\font minus \fontdimen4\font\relax}
\providecommand{\BIBforeignlanguage}[2]{{%
\expandafter\ifx\csname l@#1\endcsname\relax
\typeout{** WARNING: IEEEtran.bst: No hyphenation pattern has been}%
\typeout{** loaded for the language `#1'. Using the pattern for}%
\typeout{** the default language instead.}%
\else
\language=\csname l@#1\endcsname
\fi
#2}}
\providecommand{\BIBdecl}{\relax}
\BIBdecl

\bibitem{peng2024referring}
K.~Peng \emph{et~al.}, ``Referring atomic video action recognition,'' in \emph{ECCV}, 2024.

\bibitem{yuan2021instancerefer}
Z.~Yuan \emph{et~al.}, ``{InstanceRefer:} {Cooperative} holistic understanding for visual grounding on point clouds through instance multi-level contextual referring,'' in \emph{ICCV}, 2021.

\bibitem{wu2023referring}
D.~Wu, W.~Han, T.~Wang, X.~Dong, X.~Zhang, and J.~Shen, ``Referring multi-object tracking,'' in \emph{CVPR}, 2023.

\bibitem{seo2020urvos}
S.~Seo, J.-Y. Lee, and B.~Han, ``{URVOS:} {Unified} referring video object segmentation network with a large-scale benchmark,'' in \emph{ECCV}, 2020.

\bibitem{dang2023instructdet}
R.~Dang \emph{et~al.}, ``{InstructDET:} {Diversifying} referring object detection with generalized instructions,'' in \emph{ICLR}, 2024.

\bibitem{carreira2017quo}
J.~Carreira and A.~Zisserman, ``Quo vadis, action recognition? {A} new model and the kinetics dataset,'' in \emph{CVPR}, 2017.

\bibitem{gu2018ava}
C.~Gu \emph{et~al.}, ``{AVA:} {A} video dataset of spatio-temporally localized atomic visual actions,'' in \emph{CVPR}, 2018.

\bibitem{zhu2024vision}
L.~Zhu, B.~Liao, Q.~Zhang, X.~Wang, W.~Liu, and X.~Wang, ``Vision mamba: Efficient visual representation learning with bidirectional state space model,'' in \emph{ICML}, 2024.

\bibitem{wang2025internvideo2}
Y.~Wang \emph{et~al.}, ``{InternVideo2.5:} {Empowering} video {MLLMs} with long and rich context modeling,'' \emph{arXiv preprint arXiv:2501.12386}, 2025.

\bibitem{shi2018key}
H.~Shi, H.~Li, F.~Meng, and Q.~Wu, ``Key-word-aware network for referring expression image segmentation,'' in \emph{ECCV}, 2018.

\bibitem{ye2021referring}
L.~Ye, M.~Rochan, Z.~Liu, X.~Zhang, and Y.~Wang, ``Referring segmentation in images and videos with cross-modal self-attention network,'' \emph{TPAMI}, 2022.

\bibitem{feng2022referring}
G.~Feng, L.~Zhang, J.~Sun, Z.~Hu, and H.~Lu, ``Referring segmentation via encoder-fused cross-modal attention network,'' \emph{TPAMI}, 2023.

\bibitem{deruyttere2019talk2car}
T.~Deruyttere, S.~Vandenhende, D.~Grujicic, L.~Van~Gool, and M.-F. Moens, ``{Talk2Car:} {Taking} control of your self-driving car,'' in \emph{EMNLP}, 2019.

\bibitem{bu2022scene}
Y.~Bu \emph{et~al.}, ``Scene-text oriented referring expression comprehension,'' \emph{TMM}, 2022.

\bibitem{shi2023unsupervised}
H.~Shi, W.~Pan, Z.~Zhao, M.~Zhang, and F.~Wu, ``Unsupervised domain adaptation for referring semantic segmentation,'' in \emph{MM}, 2023.

\bibitem{zhang2023referring}
X.~Zhang, B.~Yin, Z.~Lin, Q.~Hou, D.-P. Fan, and M.-M. Cheng, ``Referring camouflaged object detection,'' \emph{TPAMI}, 2025.

\bibitem{khoreva2019video}
A.~Khoreva, A.~Rohrbach, and B.~Schiele, ``Video object segmentation with language referring expressions,'' in \emph{ACCV}, 2019.

\bibitem{jin2026interrvos}
W.~Jin, S.~Kim, J.~Lee, and S.~Kim, ``{InterRVOS:} {Interaction-aware} referring video object segmentation,'' in \emph{CVPR}, 2026.

\bibitem{lin2024echotrack}
J.~Lin \emph{et~al.}, ``{EchoTrack:} {Auditory} referring multi-object tracking for autonomous driving,'' \emph{T-ITS}, 2024.

\bibitem{wu2022referformer}
J.~Wu, Y.~Jiang, P.~Sun, Z.~Yuan, and P.~Luo, ``Language as queries for referring video object segmentation,'' in \emph{CVPR}, 2022.

\bibitem{yang2022tubedetr}
A.~Yang, A.~Miech, J.~Sivic, I.~Laptev, and C.~Schmid, ``{TubeDETR:} {Spatio-temporal} video grounding with transformers,'' in \emph{CVPR}, 2022.

\bibitem{gavrilyuk2018actor}
K.~Gavrilyuk, A.~Ghodrati, Z.~Li, and C.~G.~M. Snoek, ``Actor and action video segmentation from a sentence,'' in \emph{CVPR}, 2018.

\bibitem{mcintosh2020visual}
B.~McIntosh, K.~Duarte, Y.~S. Rawat, and M.~Shah, ``Visual-textual capsule routing for text-based video segmentation,'' in \emph{CVPR}, 2020.

\bibitem{su2023sequence}
Y.~Su, W.~Wang, J.~Liu, S.~Ma, and X.~Yang, ``Sequence as a whole: A unified framework for video action localization with long-range text query,'' \emph{TIP}, 2023.

\bibitem{yu2016modeling}
L.~Yu, P.~Poirson, S.~Yang, A.~C. Berg, and T.~L. Berg, ``Modeling context in referring expressions,'' in \emph{ECCV}, 2016.

\bibitem{chen2019weakly}
Z.~Chen, L.~Ma, W.~Luo, and K.-Y.~K. Wong, ``Weakly-supervised spatio-temporally grounding natural sentence in video,'' \emph{arXiv preprint arXiv:1906.02549}, 2019.

\bibitem{zeng2022motr}
F.~Zeng, B.~Dong, Y.~Zhang, T.~Wang, X.~Zhang, and Y.~Wei, ``{MOTR:} {End-to-end} multiple-object tracking with transformer,'' in \emph{ECCV}, 2022.

\bibitem{chung2021haa500}
J.~Chung, C.-h. Wuu, H.-r. Yang, Y.-W. Tai, and C.-K. Tang, ``{HAA500:} {Human-centric} atomic action dataset with curated videos,'' in \emph{ICCV}, 2021.

\bibitem{feichtenhofer2019slowfast}
C.~Feichtenhofer, H.~Fan, J.~Malik, and K.~He, ``{SlowFast} networks for video recognition,'' in \emph{ICCV}, 2019.

\bibitem{feichtenhofer2020x3d}
C.~Feichtenhofer, ``{X3D:} {Expanding} architectures for efficient video recognition,'' in \emph{CVPR}, 2020.

\bibitem{li2022mvitv2}
Y.~Li \emph{et~al.}, ``{MViTv2:} {Improved} multiscale vision transformers for classification and detection,'' in \emph{CVPR}, 2022.

\bibitem{ryali2023hiera}
C.~Ryali \emph{et~al.}, ``Hiera: A hierarchical vision transformer without the bells-and-whistles,'' in \emph{ICML}, 2023.

\bibitem{wang2023videomae}
L.~Wang \emph{et~al.}, ``{VideoMAE V2:} {Scaling} video masked autoencoders with dual masking,'' in \emph{CVPR}, 2023.

\bibitem{gritsenko2023end}
A.~Gritsenko \emph{et~al.}, ``End-to-end spatio-temporal action localisation with video transformers,'' \emph{arXiv preprint arXiv:2304.12160}, 2023.

\bibitem{pramono2021spatial}
R.~R.~A. Pramono, Y.-T. Chen, and W.-H. Fang, ``Spatial-temporal action localization with hierarchical self-attention,'' \emph{TMM}, 2021.

\bibitem{ji2019context}
Y.~Ji, Y.~Zhan, Y.~Yang, X.~Xu, F.~Shen, and H.~T. Shen, ``A context knowledge map guided coarse-to-fine action recognition,'' \emph{TIP}, 2020.

\bibitem{rajasegaran2023benefits}
J.~Rajasegaran, G.~Pavlakos, A.~Kanazawa, C.~Feichtenhofer, and J.~Malik, ``On the benefits of {3D} pose and tracking for human action recognition,'' in \emph{CVPR}, 2023.

\bibitem{wang2023stal}
S.~Wang, R.~Yan, P.~Huang, G.~Dai, Y.~Song, and X.~Shu, ``{Com-STAL:} {Compositional} spatio-temporal action localization,'' \emph{TCSVT}, 2023.

\bibitem{radford2021learning}
A.~Radford \emph{et~al.}, ``Learning transferable visual models from natural language supervision,'' in \emph{ICML}, 2021.

\bibitem{li2022blip}
J.~Li, D.~Li, C.~Xiong, and S.~Hoi, ``{BLIP:} {Bootstrapping} language-image pre-training for unified vision-language understanding and generation,'' in \emph{ICML}, 2022.

\bibitem{li2023blip}
J.~Li, D.~Li, S.~Savarese, and S.~Hoi, ``{BLIP-2:} {Bootstrapping} language-image pre-training with frozen image encoders and large language models,'' in \emph{ICML}, 2023.

\bibitem{luo2022clip4clip}
H.~Luo \emph{et~al.}, ``{CLIP4Clip:} {An} empirical study of {CLIP} for end to end video clip retrieval and captioning,'' \emph{Neurocomputing}, 2022.

\bibitem{ma2022x}
Y.~Ma, G.~Xu, X.~Sun, M.~Yan, J.~Zhang, and R.~Ji, ``{X-CLIP:} {End-to-end} multi-grained contrastive learning for video-text retrieval,'' in \emph{MM}, 2022.

\bibitem{wang2023actionclip}
M.~Wang, J.~Xing, J.~Mei, Y.~Liu, and Y.~Jiang, ``{ActionCLIP:} {Adapting} language-image pretrained models for video action recognition,'' \emph{TNNLS}, 2023.

\bibitem{zhang2023multi}
G.~Zhang, J.~Ren, J.~Gu, and V.~Tresp, ``Multi-event video-text retrieval,'' in \emph{CVPR}, 2023.

\bibitem{wu2023cap4video}
W.~Wu, H.~Luo, B.~Fang, J.~Wang, and W.~Ouyang, ``{Cap4Video:} {What} can auxiliary captions do for text-video retrieval?'' in \emph{CVPR}, 2023.

\bibitem{chen2023tagging}
Y.~Chen, J.~Wang, L.~Lin, Z.~Qi, J.~Ma, and Y.~Shan, ``Tagging before alignment: Integrating multi-modal tags for video-text retrieval,'' in \emph{AAAI}, 2023.

\bibitem{wang2022internvideo}
Y.~Wang \emph{et~al.}, ``{InternVideo:} {General} video foundation models via generative and discriminative learning,'' \emph{arXiv preprint arXiv:2212.03191}, 2022.

\bibitem{yang2022learning}
A.~Yang, A.~Miech, J.~Sivic, I.~Laptev, and C.~Schmid, ``Learning to answer visual questions from web videos,'' \emph{TPAMI}, 2025.

\bibitem{xiao2023contrastive}
J.~Xiao \emph{et~al.}, ``Contrastive video question answering via video graph transformer,'' \emph{TPAMI}, 2023.

\bibitem{gao2023mist}
D.~Gao, L.~Zhou, L.~Ji, L.~Zhu, Y.~Yang, and M.~Z. Shou, ``{MIST:} {Multi-modal} iterative spatial-temporal transformer for long-form video question answering,'' in \emph{CVPR}, 2023.

\bibitem{li2022representation}
J.~Li, L.~Niu, and L.~Zhang, ``From representation to reasoning: Towards both evidence and commonsense reasoning for video question-answering,'' in \emph{CVPR}, 2022.

\bibitem{gandhi2022measuring}
M.~Gandhi, M.~O. Gul, E.~Prakash, M.~Grunde-McLaughlin, R.~Krishna, and M.~Agrawala, ``Measuring compositional consistency for video question answering,'' in \emph{CVPR}, 2022.

\bibitem{li2023lavender}
L.~Li \emph{et~al.}, ``{LAVENDER:} {Unifying} video-language understanding as masked language modeling,'' in \emph{CVPR}, 2023.

\bibitem{bagad2023test}
P.~Bagad, M.~Tapaswi, and C.~G.~M. Snoek, ``Test of time: Instilling video-language models with a sense of time,'' in \emph{CVPR}, 2023.

\bibitem{2023videochat}
K.~Li \emph{et~al.}, ``{VideoChat:} {Chat-centric} video understanding,'' \emph{Science China Information Sciences}, 2025.

\bibitem{zhang2023video}
H.~Zhang, X.~Li, and L.~Bing, ``{Video-LLaMA:} {An} instruction-tuned audio-visual language model for video understanding,'' in \emph{EMNLP}, 2023.

\bibitem{ahmed2025qwen}
I.~Ahmed, S.~Islam, P.~P. Datta, I.~Kabir, M.~N. U.~R. Chowdhury, and A.~Haque, ``{Qwen} 2.5: {A} comprehensive review of the leading resource-efficient {LLM} with potentioal to surpass all competitors,'' 2025.

\bibitem{ryoo2021tokenlearner}
M.~S. Ryoo, A.~Piergiovanni, A.~Arnab, M.~Dehghani, and A.~Angelova, ``{TokenLearner:} {What} can 8 learned tokens do for images and videos?'' \emph{arXiv preprint arXiv:2106.11297}, 2021.

\bibitem{rao2021dynamicvit}
Y.~Rao, W.~Zhao, B.~Liu, J.~Lu, J.~Zhou, and C.-J. Hsieh, ``{DynamicViT:} {Efficient} vision transformers with dynamic token sparsification,'' in \emph{NeurIPS}, 2021.

\bibitem{liu2021cross}
S.~Liu, T.~Hui, S.~Huang, Y.~Wei, B.~Li, and G.~Li, ``Cross-modal progressive comprehension for referring segmentation,'' \emph{TPAMI}, 2022.

\bibitem{nagrani2021attention}
A.~Nagrani, S.~Yang, A.~Arnab, A.~Jansen, C.~Schmid, and C.~Sun, ``Attention bottlenecks for multimodal fusion,'' in \emph{NeuIPS}, 2021.

\bibitem{le2020hierarchical}
T.~M. Le, V.~Le, S.~Venkatesh, and T.~Tran, ``Hierarchical conditional relation networks for video question answering,'' in \emph{CVPR}, 2020.

\bibitem{jiang2020divide}
J.~Jiang, Z.~Chen, H.~Lin, X.~Zhao, and Y.~Gao, ``Divide and conquer: Question-guided spatio-temporal contextual attention for video question answering,'' in \emph{AAAI}, 2020.

\bibitem{zhang2022opt}
S.~Zhang \emph{et~al.}, ``{OPT:} {Open} pre-trained transformer language models,'' \emph{arXiv preprint arXiv:2205.01068}, 2022.

\bibitem{dosovitskiy2020vit}
A.~Dosovitskiy \emph{et~al.}, ``An image is worth 16x16 words: Transformers for image recognition at scale,'' in \emph{ICLR}, 2021.

\bibitem{carion2020end}
N.~Carion, F.~Massa, G.~Synnaeve, N.~Usunier, A.~Kirillov, and S.~Zagoruyko, ``End-to-end object detection with transformers,'' in \emph{ECCV}, 2020.

\bibitem{kingma2014adam}
D.~P. Kingma and J.~Ba, ``Adam: A method for stochastic optimization,'' in \emph{ICLR}, 2015.

\bibitem{lei2022revealing}
J.~Lei, T.~L. Berg, and M.~Bansal, ``Revealing single frame bias for video-and-language learning,'' \emph{ACL}, 2023.

\bibitem{kirillov2023segment}
A.~Kirillov \emph{et~al.}, ``Segment anything,'' in \emph{ICCV}, 2023.

\bibitem{jin2023refclip}
L.~Jin \emph{et~al.}, ``{RefCLIP:} {A} universal teacher for weakly supervised referring expression comprehension,'' in \emph{CVPR}, 2023.

\bibitem{zong2023mcomet}
D.~Zong and S.~Sun, ``{McOmet:} {Multimodal} fusion transformer for physical audiovisual commonsense reasoning,'' in \emph{AAAI}, 2023.

\bibitem{lin2017focal}
T.-Y. Lin, P.~Goyal, R.~Girshick, K.~He, and P.~Doll{\'a}r, ``Focal loss for dense object detection,'' in \emph{ICCV}, 2017.

\bibitem{openai2022chatgpt}
OpenAI, ``{ChatGPT:} {Optimizing} language models for dialogue,'' \url{https://openai.com/}, 2022.

\end{thebibliography}

\end{document}